\documentclass[final,5p,times,twocolumn,numbering]{elsarticle}

\usepackage{subcaption}
\usepackage{graphicx}
\usepackage{amssymb}
\usepackage{hyperref}
\usepackage{float}
\usepackage{pbox}
\usepackage{lipsum}
\usepackage{xcolor}
\usepackage{lineno}
\usepackage{soul}
\usepackage{amsmath}
\usepackage{amssymb}

\usepackage{multirow}
\usepackage{makecell}

\usepackage{gensymb}

\begin{document}
	\begin{frontmatter}
		
		\title{Selective Cotton Boll Localization for Robotic Harvesting: Evaluation of Deep Learning Vision Models Under Field Conditions}

		\author[a]{Thevathayarajh Thayananthan (\href{mailto:theva@uga.edu}{theva@uga.edu})}
		\author[a]{Xin Zhang* (\href{mailto:Xin.Zhang2@uga.edu}{Xin.Zhang2@uga.edu})}
		\author[a]{Isuru Laddusinghe Badu}
		\author[a]{Jonathan Harjono}
		\author[b]{Glen C. Rains}
		\author[a]{Beiwen Li}
		\author[c]{Leonardo M. Bastos}
		\author[d]{Nuwan K. Wijewardane}
		\author[d]{Vitor S. Martins}

		\affiliation[a]{organization={School of Environmental, Civil, Agricultural and Mechanical Engineering, University of Georgia},
			city={Athens},
			postcode={30602}, 
			state={GA},
			country={USA}}
		
		\affiliation[b]{organization= {Department of Entomology, University of Georgia},
			city={Tifton},
			postcode={31793}, 
			state={GA},
			country={USA}}
		
		\affiliation[c]{organization={Department of Crop and Soil Sciences, University of Georgia},
			city={Athens},
			postcode={30602}, 
			state={GA},
			country={USA}}
			
		\affiliation[d]{organization={Department of Agricultural and Biological Engineering, Mississippi State University},
            city={Mississippi State},
            postcode={39762}, 
            state={MS},
            country={USA}}

		\begin{abstract}
			The United States is one of the world’s leading cotton producers, with production concentrated across the cotton belt in 17 southern states. Conventional cotton harvesting relies on large, heavy mechanical pickers that are typically deployed only once at the end of the season following chemical defoliation. This practice introduces several challenges, including soil compaction, reduced cotton quality due to delayed harvesting of early-opened bolls, cotton loss, and high operational costs.
			\textcolor{black}{A lightweight, autonomous, vision-guided robotic cotton picker could help address these limitations by selectively harvesting cotton bolls as they reach peak quality throughout the maturation period. Reliable cotton boll detection and segmentation under natural field conditions are essential requirements for such a robotic system. This study developed and evaluated a deep-learning-based perception framework for selective robotic cotton picking. The dataset contained 1,008 annotated field images collected using three cameras under varying natural lighting and weather conditions. Object-detection models from the YOLOv8 through YOLOv13 families were evaluated using their default configurations, while segmentation performance was assessed using YOLOv8-seg, YOLOv11-seg, YOLOv12-seg, the Segment Anything Model (SAM), SAMv2.1, FastSAM, and Grounded-SAM with the Recognize Anything Model (RAM). Model performance was evaluated based on accuracy--speed trade-offs, segmentation-area consistency, and robustness across five randomized dataset splits.}
			Among the detection models, GELAN-s achieved the most favorable balance between mean average precision (mAP) and inference speed, obtaining an mAP of 86.1\%, precision of 81.6\%, recall of 76.6\%, and an F1-score of 79.0\%, with an average inference time of 42.3~ms per image.
			Among the direct segmentation models, YOLOv12-m-seg provided the most favorable \textcolor{black}{balance between AP@0.5 and FPS}, achieving a \textcolor{black}{segmentation AP@0.5} of 83.7\% with an inference time of 20.4~ms per image. \textcolor{black}{In the detection-prompted segmentation approach, bounding-box prompts generated by GELAN-s improved the localization of cotton bolls for SAM and SAMv2.1, while SAMv2.1 Tiny consistently outperformed FastSAM and Grounded-SAM with RAM. In the area-based evaluation against manually annotated segmentation masks, YOLOv12-m-seg achieved an $R^2$ value of 0.966, compared with 0.860 for GELAN-s + SAMv2.1 Tiny. Although YOLOv12-m-seg produced less-negative count-based $R^2$ values than GELAN-s, all count-based values remained negative, indicating poor absolute agreement with the reference cotton boll counts.}
			Field experiments conducted using a UR5e robotic manipulator, a custom end-effector, and a ZED2i stereo camera further validated the effectiveness of the YOLOv12-m-seg model for real-time cotton boll detection, segmentation, and selective picking under varying confidence levels.
			\textcolor{black}{These results demonstrate that YOLOv12-m-seg provides an efficient perception model for robotic cotton harvesting and has strong potential for field deployment. The complete image dataset, annotations, and trained weights are publicly available at \url{https://github.com/imtheva/CottonBoll_Harvest}.}
		\end{abstract}

		
	\end{frontmatter}
	
	\section*{\textcolor{black}{HIGHLIGHTS}}
	
	\begin{itemize}
		
		\item \textcolor{black}{A total of 33 YOLO detectors were benchmarked on 1,008 cotton field images.}
		
		\item \textcolor{black}{GELAN-s achieved 86.1\% mAP@0.5 with 42.3 ms inference per image.}
		
		\item \textcolor{black}{YOLOv12-m-seg achieved 83.7\% mAP@0.5 with 20.4 ms inference per image.}
		
		\item \textcolor{black}{YOLOv12-m-seg achieved an area $R^2$ of 0.966 against manual masks.}
		
		\item \textcolor{black}{Field tests validated Cotton-Eye perception for robotic cotton picking.}
		
	\end{itemize}
	
\section{INTRODUCTION}\label{introduction}

    Computer vision is one of the most rapidly advancing technologies worldwide, driving automation through the analysis of visual data. As a branch of artificial intelligence (AI), computer vision aims to enable machines to perceive and interpret their surroundings by integrating techniques from image processing, machine learning, and digital technologies \cite{che2024intelligent}. \textcolor{black}{Stereo vision, a \textcolor{black}{common} computer vision technique, estimates depth by analyzing images captured from two spatially separated cameras, similar to human binocular vision.} \textcolor{black}{This capability is particularly relevant to robotic systems because it supports both target identification and the estimation of spatial information required for accurate target localization and manipulation.} \textcolor{black}{Recent advances in AI have significantly transformed computer vision and robotic perception, enabling more accurate and robust object detection in complex agricultural environments. Early computer vision systems relied on classical techniques that used manually designed features, such as edges, corners, appearance, and geometric information, for object recognition \cite{hussain2025classical}.} \textcolor{black}{Although these approaches can be effective in controlled settings, their performance may be sensitive to changes in illumination, background complexity, target occlusion, and crop variability.} \textcolor{black}{The development of machine learning (ML), a subset of AI that enables computers to learn patterns from data, laid the foundation for modern intelligent perception systems. Building upon ML, deep learning has emerged as the dominant paradigm in computer vision by employing artificial neural networks to automatically learn hierarchical features from large datasets \cite{Ultralytics2026DeepLearning}. Consequently, deep learning has become the primary perception approach in modern agricultural robotics, providing more robust and accurate object detection and localization under challenging field conditions than traditional computer vision methods. Accordingly, recent research on robotic cotton harvesting has predominantly focused on deep learning-based perception pipelines for cotton boll detection, localization, and maturity assessment.}

	In recent years, agriculture has significantly benefited from computer vision-driven automation, particularly in robotics and machinery. The applications of computer vision in agriculture span a wide range of subdomains, including crop monitoring \cite{story2015design, d2022monitoring}, weed detection and control \cite{turkouglu2019plant, liu2019pestnet,kasinathan2021machine}, harvesting \cite{zhang2023multi,benavides2020automatic,zhang2020multi}, autonomous navigation \cite{zhao2020ground,panda2023agronav}, and \textcolor{black}{Uncrewed} Aerial Vehicle (UAVs)-based \textcolor{black}{crop scouting and monitoring} \cite{gunder2022agricultural,shammi2024application}. Among these, cotton harvesting stands out as a particularly promising area for the integration of computer vision into robotic systems, offering \textcolor{black}{a} potential solution to long-standing challenges in traditional agricultural practices.

    Cotton farming is one of the major agricultural sectors in the United States and presents a key opportunity for integrating computer vision into robotic systems to automate harvesting. As one of the world’s largest cotton producers, the U.S. plays a leading role in global exports, accounting for more than 35\% of the raw cotton export market \cite{USDA_Cotton}. Currently, U.S. \textcolor{black}{cotton} farmers depend on large, expensive mechanized \textcolor{black}{cotton harvesters}, which \textcolor{black}{can cost approximately} \$1 million \cite{john_deere_cp770, allmachines_cp770}. These traditional harvesting methods require \textcolor{black}{most harvestable cotton bolls to reach maturity before harvesting}, meaning the machines are deployed only \textcolor{black}{once} at the end of the season. Additionally, chemical defoliants \textcolor{black}{are typically applied when} at least 60\% of the cotton bolls have opened \textcolor{black}{to facilitate a single large-scale harvest}. \textcolor{black}{Although this once-over approach supports high-throughput harvesting, it does not allow individual cotton bolls to be selectively harvested as they become ready.} \textcolor{black}{Moreover, this harvesting strategy creates several agronomic, economic, and environmental challenges. Because bolls at the highest-yielding fruiting positions typically open first, waiting for bolls at later-developing positions to open prolongs the exposure of these high-yielding bolls to adverse weather, which can degrade their fiber quality} \cite{barnes2021opportunities}. Additionally, the substantial weight of these machines compacts the soil, restricting root growth and limiting water and nutrient uptake, ultimately leading to yield losses \cite{gharakhani2023integration}. \textcolor{black}{Reliance on chemical defoliants further increases production costs and raises environmental concerns.} These challenges underscore the need for innovative, computer-vision-assisted harvesting solutions to enhance efficiency and sustainability in cotton production.
	
	One promising solution to the limitations of conventional harvesting is the development of lightweight, autonomous robots equipped with vision-assisted picking capabilities. These robots can help reduce soil compaction, preserve cotton quality, and reduce the reliance on chemical inputs, thereby promoting more sustainable farming practices. Unlike traditional harvesters, small autonomous robots \textcolor{black}{are expected to} navigate through fields and selectively harvest cotton bolls as they reach maturity, allowing for precise and timely picking. Within this context, computer vision is essential for identifying and localizing cotton bolls in the complex field environments, enabling robotic manipulators to perform accurate and efficient harvesting operations.

	Several studies have investigated the application of computer vision techniques for detecting different maturity stages of cotton bolls using UAVs, ground-based robots, and stereo camera imagery \cite{liu2023small, yeom2018automated, singh2021image, sun2019image, zhang2024yolo, gharakhani2024field, nagarajan2023cotton}. One such study utilized the MRF-YOLO \textcolor{black}{(Multi-Receptive Field–You Only Look Once)} model, based on YOLOX \cite{ge2021yolox}, in combination with a \textcolor{black}{Multi-Receptive Field (MRF)} extraction technique to enhance detection accuracy for unopened cotton bolls \textcolor{black}{to understand the growth information and for cotton field management} \cite{liu2023small}. To further improve \textcolor{black}{the detection performance of} small cotton boll, a multi-scale residual block and an attention module were integrated into the model. Additionally, the mixed-up operation was removed, and the mosaic operation was disabled during the last 15 training epochs to increase detection accuracy. As a result, MRF-YOLO demonstrated strong performance in detecting and counting unopened cotton bolls, even in obscured environments, achieving mean Average Precision (mAP), precision, and recall values of 92.75\%, 92.61\%, and 90.06\%, respectively.

	In \textcolor{black}{a related} study, ultra-fine spatial resolution UAV images were captured, and random seed points were extracted from selected image subsets \cite{yeom2018automated}. A region-growing algorithm was then applied to determine whether the generated segments corresponded to cotton bolls. The Otsu method \cite{otsu1975threshold} was used to distinguish cotton bolls from other candidates based on spectral information and brightness threshold values. Finally, a binary classification approach was applied using the derived threshold values to detect cotton bolls. The method achieved an accuracy of 88\%, successfully identifying cotton boll \textcolor{black}{classification} that were directly related to actual yield.
	
	In \textcolor{black}{another work}, the YOLO SSPD (Small-Scale Pyramid Depth-Aware Detection) model was used to detect small cotton bolls, even with the low resolution of UAV imagery \cite{zhang2024yolo}. The YOLOv8 \cite{yolov8_ultralytics}  architecture was employed for transfer learning, integrating space-to-depth and non-strided convolution (SPD-Conv) with a specialized small-target detection head. Additionally, a parameter-free attentional mechanism was incorporated, significantly enhancing target boll detection accuracy. These advancements, combined with transfer learning techniques applied to UAV imagery, resulted in a \textcolor{black}{detection precision} accuracy of 87.4\% and a cotton boll count correlation ($R^2$) of 0.86, demonstrating improved detection performance for small cotton bolls.

	\textcolor{black}{Singh et al. \cite{singh2021image}} explored the performance of four image processing algorithms for segmenting cotton bolls under outdoor field conditions. The \textcolor{black}{group} evaluated four color models (\textcolor{black}{i.e}, RGB normalized RGB (sRGB), HSV, and YCbCr) to determine the most effective model for cotton field segmentation. Three of the image processing algorithms such as color difference, band ratio, and chromatic aberration \textcolor{black}{used} the RGB color \textcolor{black}{model}, while the fourth algorithm utilized the YCbCr color model. Among \textcolor{black}{all}, the chromatic aberration method demonstrated the highest identification rate of 91.05\% and exhibited lower error rates compared to the other approaches. Despite its strong performance, the method encountered challenges such as false positive detection due to light reflections \textcolor{black}{from bolls} and difficulty distinguishing overlapping bolls.

	\textcolor{black}{A} traditional image-processing technique \textcolor{black}{was also employed} to address challenges associated \textcolor{black}{with} cotton boll counting \cite{sun2019image}. Cotton bolls were segmented from the background using a double-threshold method combined with a region-growing algorithm that incorporated both color and spatial features. Additionally, boll counts were estimated using three geometric feature-based algorithms. To improve segmentation accuracy, disjointed regions caused by branches and burrs \textcolor{black}{(dry, woody husk of the cotton boll)} were merged using line features extracted through the linear Hough Transform and boundary distance measurements.  Overlapping bolls were further distinguished based on area and elongation ratio. The proposed segmentation and counting method achieved an accuracy of 83\% \textcolor{black}{in cotton boll count}, demonstrating effectiveness under complex field conditions.
	
	\textcolor{black}{Furthermore, a} vision-guided robotic harvester prototype was developed and tested for cotton picking \cite{gharakhani2024field}. The system utilized a three-fingered end-effector, a \textcolor{black}{stereo vision} camera, and a 3-degree-of-freedom (DOF) linear robotic arm. The YOLOv4-tiny algorithm was implemented to detect and localize cotton bolls based on the visual input from the stereo camera. To reduce light interference and enhance detection accuracy, a black background was placed behind the plants.  The \textcolor{black}{robotic} system achieved detection, localization, and picking performances of 78.1\%, 70.0\%, and 83.1\%, respectively. Additionally, the robotic picker was able to harvest 55.1\% of the seed cotton that was both visible and within the arm’s workspace. The system’s performance could be further improved by refining the detection or segmentation approach, enhancing localization accuracy, and optimizing the robot’s control procedures.
	
	\textcolor{black}{A recent cotton segmentation study employed the convolutional neural network (CNN)-based U-Net deep learning model to address challenges associated with sky interference in outdoor cotton boll images \cite{nagarajan2023cotton}.} The performance of U-Net was compared against the VGG16 model \cite{simonyan2014very} for cotton boll segmentation. After 1,000 training epochs, the U-Net model achieved a validation accuracy of 99.0\%, outperforming VGG16, which achieved an accuracy of 94.5\%.
	
	However, most existing studies focus on cotton boll detection or segmentation using either traditional image processing or deep learning methods, each presenting specific limitations. One of the critical factors for robotic picking is near real-time detection \textcolor{black}{and/or segmentation} , which is essential for sending command controls immediately after localizing the cotton boll. Many of the reviewed studies do not emphasize inference time, which is crucial for \textcolor{black}{field deployment}. Additionally, these studies either focus on detecting all cotton bolls in an image or segmenting them without considering real-time constraints. While the use of a solid or opaque background can enhance cotton boll detection, it is impractical for \textcolor{black}{in-field} implementation in the field. Simple detection and segmentation approaches may provide real-time performance but often lack precise segmentation capabilities \textcolor{black}{for harvesting robots}, leading to inaccuracies in occluded or clustered bolls.
	
	Therefore, a reliable cotton boll localization approach is \textcolor{black}{urgently needed} for a mobile autonomous cotton picker to accurately identify seed cotton for harvesting. While YOLO-based object detection algorithms have demonstrated promising results in cotton boll identification, they primarily generate bounding boxes rather than precise, pixel-level segmentations \textcolor{black}{which is} an important limitation for accurate, targeted robotic picking. 
	
	This study aims to develop a cotton boll detection and localization system for a robotic cotton picker operating in cotton fields. Specifically, the objectives of this study are to: (1) compare the performance of recent YOLO algorithms for seed cotton boll detection and segmentation using their default hyperparameters and configurations; (2) implement and analyze segmentation using Segment Anything Model (SAM) variants prompted by YOLO detections and native YOLO segmentation models; (3) compare the segmentation performance of SAM-based approaches with FastSAM, Grounded-SAM, and YOLO segmentation models; and (4) assess the performance of cotton boll detection and segmentation for localizing medium and large cotton bolls located on the camera-facing side of the plant canopy and within the accessible workspace of the robotic manipulator under outdoor field conditions and operational constraints relevant to robotic harvesting.
	
	\section{MATERIALS AND METHODS}
	
	\subsection{Cotton-Eye Perception System}\label{sec:cottoneye}

	\begin{figure}[hbt!]
		\centering
		\begin{subfigure}[b]{0.49\textwidth}
			\centering
			\includegraphics[height=4.0 cm]{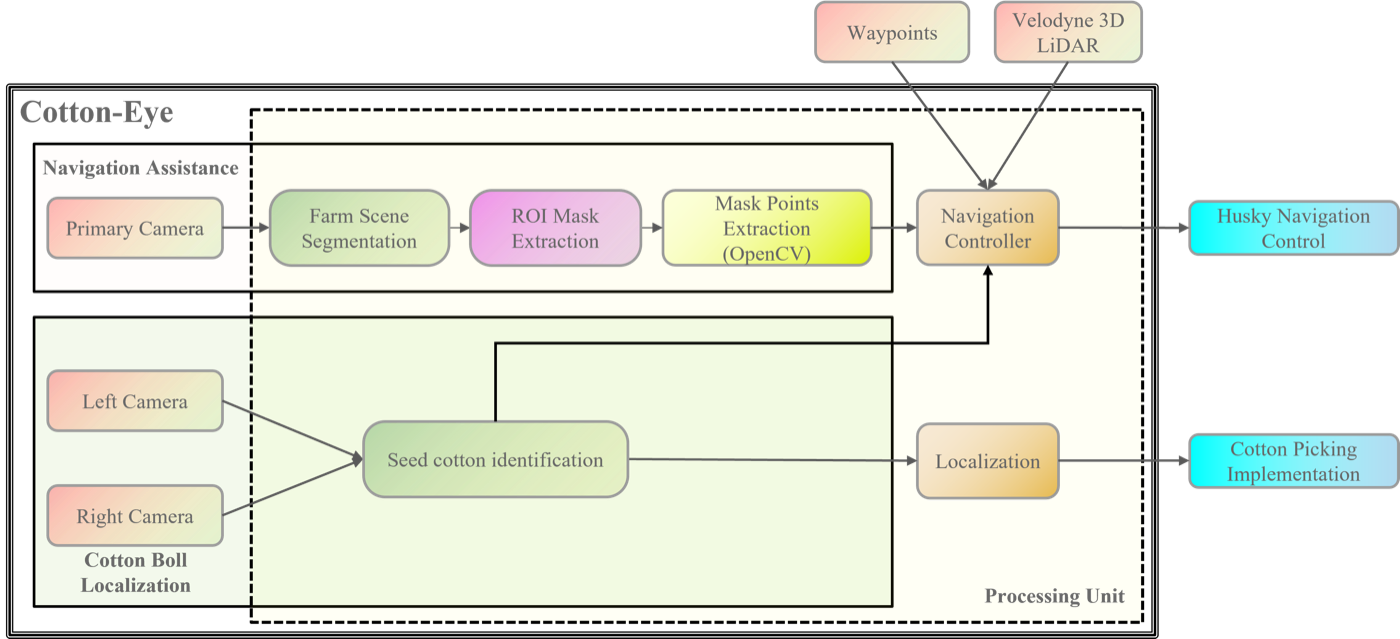}
		\end{subfigure}
		\caption{Overview of the Cotton-Eye perception system integrated \textcolor{black}{with} the autonomous cotton picker, containing two modules: Navigation Assistance and Cotton Boll Localization.}
		\label{figure:1}    
	\end{figure}
	
	This study was conducted as part of the Cotton-Eye perception system (Figure \ref{figure:1}), which serves as the computer vision-based system integrated into an autonomous cotton picker for navigation and cotton boll picking. The Cotton-Eye system comprises two primary components: \textcolor{black}{Navigation Assistance} and \textcolor{black}{Cotton Boll Localization}, both relying on visual input from three strategically \textcolor{black}{installed} cameras.

	The Navigation Assistance component supports the autonomous cotton picker by aiding its navigation through visual perception-based guidance, supplemented by additional sensors such as an IMU \textcolor{black}{Inertial Measurement Unit}, 3D LiDAR \textcolor{black}{Light Detection and Ranging}, and a Vision-RTK \textcolor{black}{Real-Time Kinematic} GPS \textcolor{black}{Global Positioning System}. Among the three cameras integrated \textcolor{black}{with} the Cotton-Eye system, the primary front-facing camera is responsible for farm scene segmentation and mask extraction. The extracted mask-based coordinates \textcolor{black}{are used to} assist the cotton picker for navigation.
	
	Meanwhile, the Cotton Boll Localization component utilizes the remaining two cameras, mounted on both sides of the cotton picker, to detect and segment cotton for harvesting. This study focuses on developing the Cotton Boll Localization component of the Cotton-Eye \textcolor{black}{perception} system to improve the precision and efficiency of cotton boll localization. 
	\textcolor{black}{The complete methodological workflow is summarized in Figure~\ref{Fig-0-workflow}, while the individual stages are described in detail in the subsequent sections.}
	
	\subsection{\textcolor{black}{High-Level Methodological Framework}}
	\label{sec:framework}
	
	\begin{figure*}[t]
		\centering
		\includegraphics[width=\textwidth]{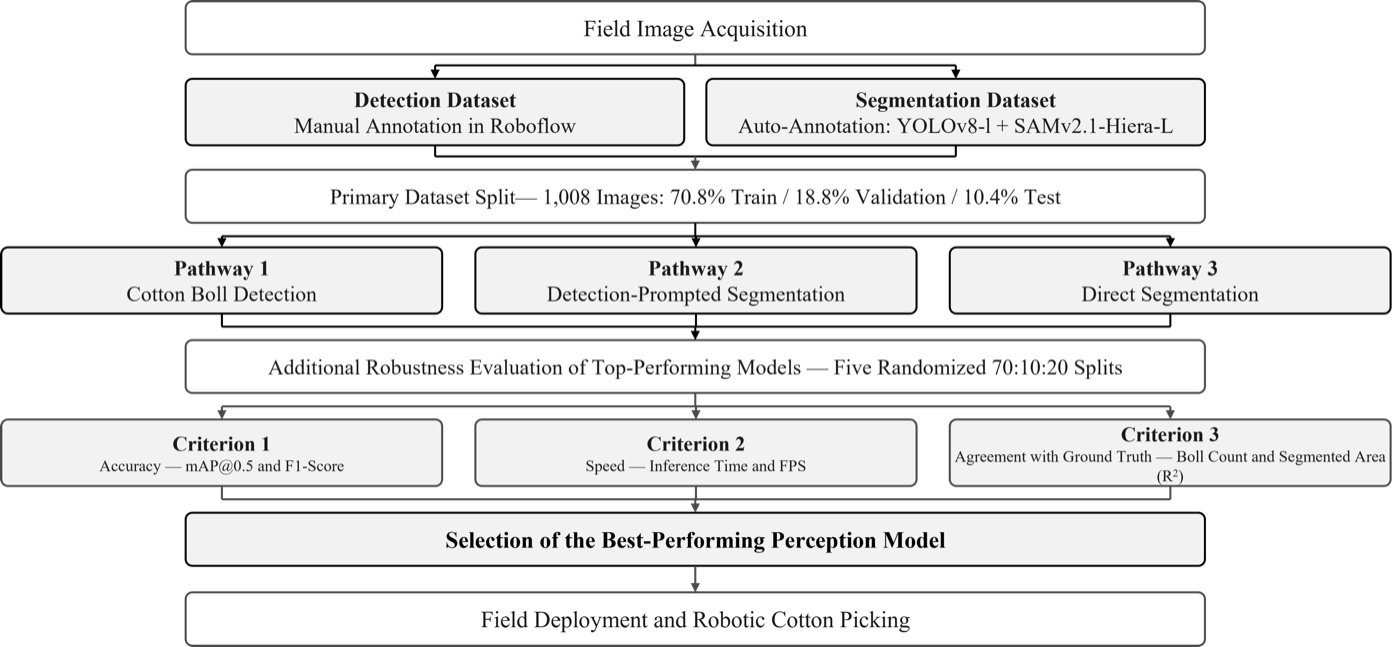}
		\caption{\textcolor{black}{High-level overview of the proposed cotton boll perception framework, including dataset preparation, the three model-evaluation pathways, robustness analysis, model-selection criteria, and field deployment for robotic cotton picking.}}
		\label{Fig-0-workflow}
	\end{figure*}
	
	\textcolor{black}{As illustrated in Figure~\ref{Fig-0-workflow}, the field images were used to construct corresponding detection and segmentation datasets. The detection dataset was manually annotated with cotton boll bounding boxes using Roboflow (Roboflow Inc., Des Moines, IA, USA), whereas the segmentation dataset was automatically annotated using YOLOv8-l and SAMv2.1-Hiera-L. Both datasets contained the same 1,008 field images and followed the primary split of 70.8\% for training, 18.8\% for validation, and 10.4\% for testing.}
	
	\textcolor{black}{The study evaluated three complementary perception pathways: (1) cotton boll detection using bounding-box-based models, (2) detection-prompted segmentation using bounding boxes generated by the selected detection model as prompts for SAM-based models, and (3) direct pixel-level segmentation using YOLO-based segmentation models.} 

    \textcolor{black}{The models were initially evaluated using the primary dataset split based on detection or segmentation accuracy and inference speed. To reduce dependence on this single train–validation–test partition, the top-performing models were subsequently retrained and evaluated using five randomized 70:10:20 splits. Performance was summarized across the five runs to assess sensitivity to the assignment of images among the training, validation, and test subsets and to provide a descriptive assessment of model consistency across alternative partitions of the same dataset.}
		
	Final model selection considered three criteria: accuracy based on mAP@0.5 and F1 score, computational efficiency based on inference time and frames per second (FPS), and \textcolor{black}{agreement assessed using cotton boll count and segmented area analyses based on the coefficient of determination ($R^2$). Cotton boll counts were compared with reference counts derived from the automatically annotated test masks. Segmented areas were compared with the manually annotated masks prepared for the 105 test images.} The best performing perception model identified through these evaluations was subsequently integrated into the Cotton Eye system for field deployment and robotic cotton picking experiments.

	\subsection{Imagery Acquisition}\label{sec:imageacquisition}
	
	\begin{figure}[hbt!]
		\centering
		\begin{subfigure}[b]{0.49\textwidth}
			\centering
			\includegraphics[height=5cm]{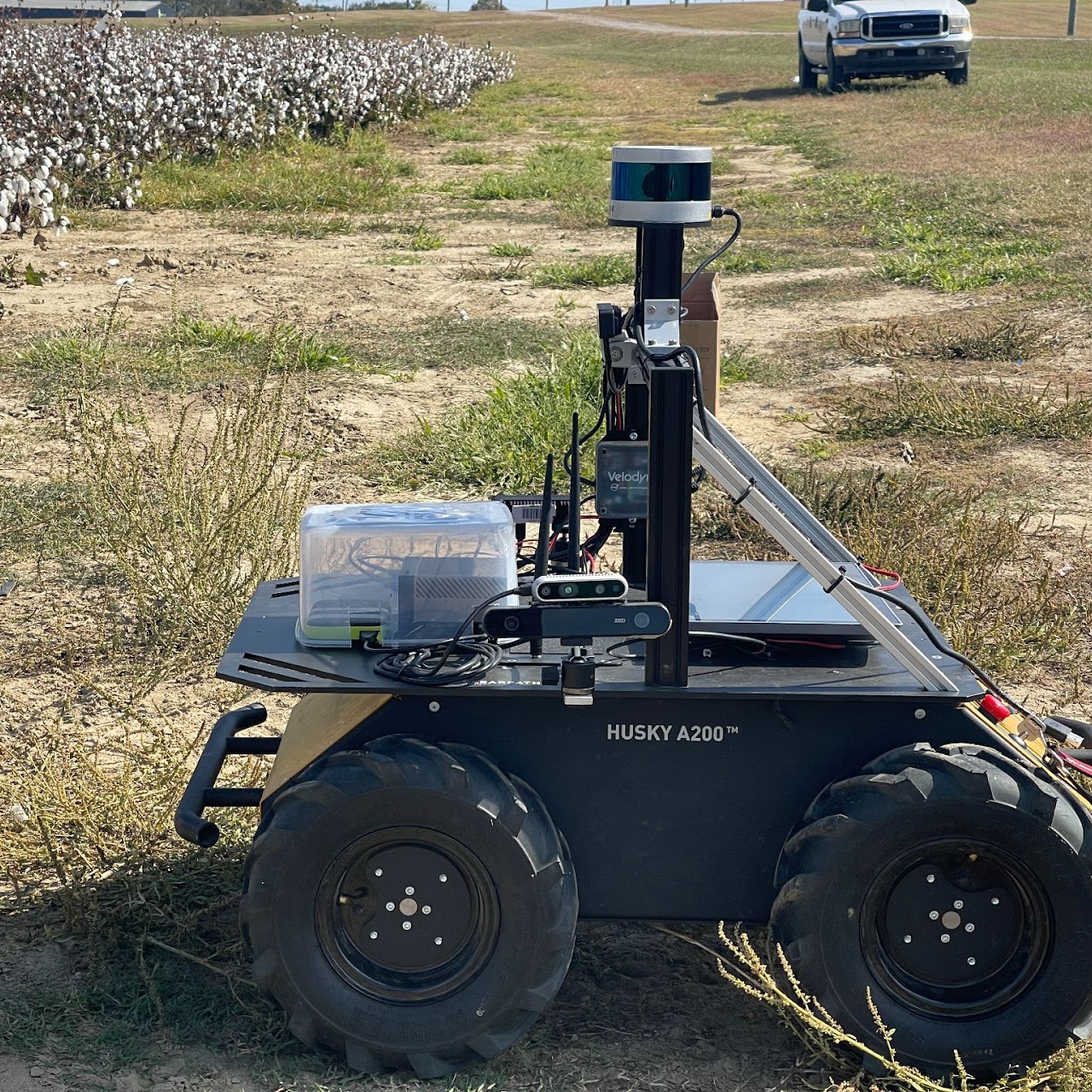}
		\end{subfigure}
		\caption{Clearpath's Husky entering the cotton farm to capture images\textcolor{black}{.}}
		\label{figure:2}    
	\end{figure}

	The imagery datasets were collected using a Clearpath Husky \textcolor{black}{A200} (Clearpath Robotics, Ontario, Canada; parent company: Rockwell Automation, Inc., Milwaukee, WI, USA) platform equipped with four cameras, as shown in Figure \ref{figure:2}: two OAK-D Pro (Luxonis, Littleton, CO, USA), one \textcolor{black}{ZED2i} (Stereolabs, San Francisco, CA, USA), and one RealSense D435i (Intel Services Division LLC, Santa Clara, CA, USA). One OAK-D Pro camera was mounted at the front of the robot, while the \textcolor{black}{other} was positioned on the right side. The RealSense D435i and \textcolor{black}{ZED2i} cameras were installed on the left side to capture images. However, only three \textcolor{black}{side-facing cameras} were used for the cotton detection and segmentation approach. The front-facing OAK-D Pro camera was designated for visual-\textcolor{black}{guided} autonomous navigation \textcolor{black}{in the future}.
	
	\begin{figure}[hbt!]
		\centering
		\begin{subfigure}[b]{0.49\textwidth}
			\centering
			\includegraphics[height=5cm]{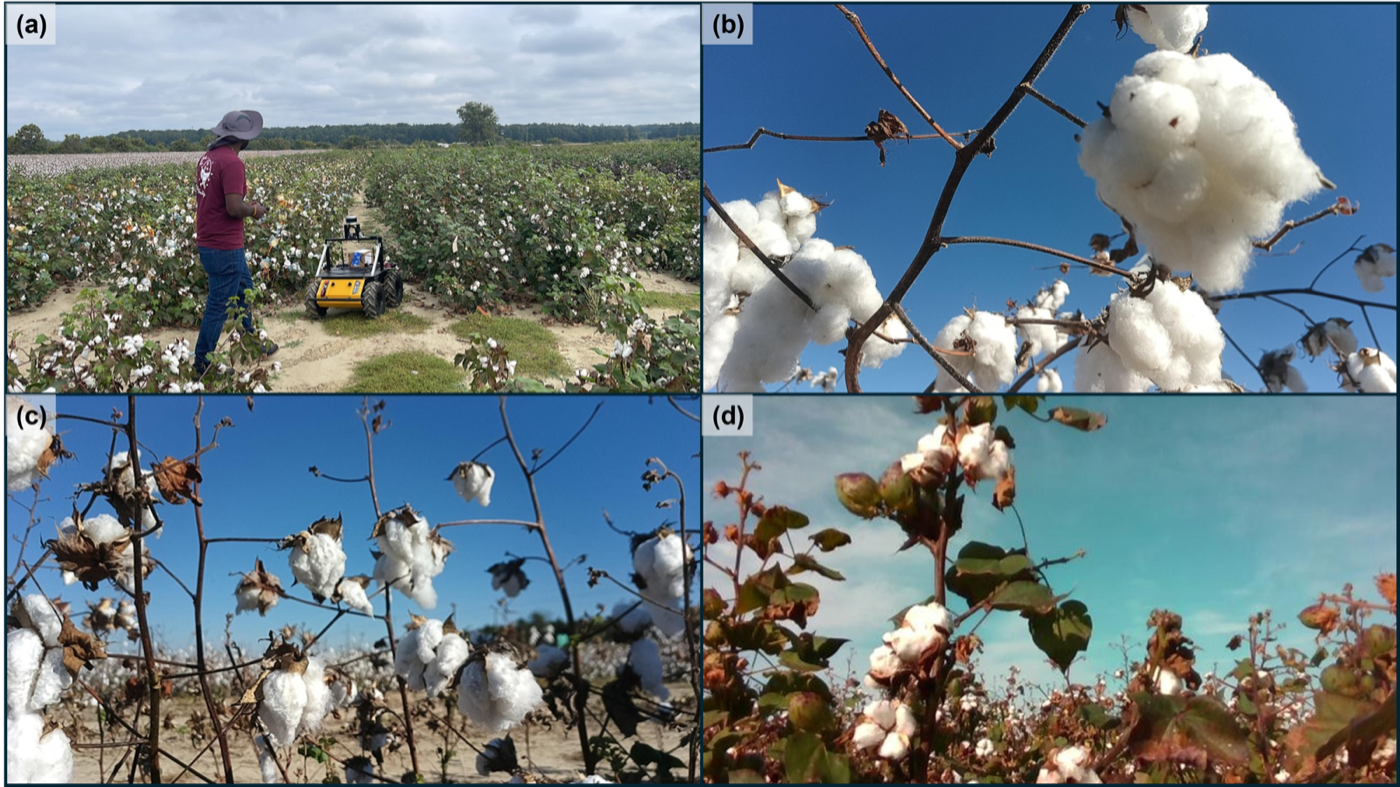}
		\end{subfigure}
		\caption{\textcolor{black}{Image} data acquisition (a) \textcolor{black}{at the North Farm Research} field, Mississippi State University using (b) \textcolor{black}{a right-facing} OAK-D Pro; (c) \textcolor{black}{a left-facing ZED2i}; (d) \textcolor{black}{a left-facing} RealSense D435i.}
		\label{figure:3}    
	\end{figure}
	
	Data collection \textcolor{black}{was conducted} from September 28, 2023, to November 8, 2023, at the R. R. Foil Plant Science Research Center (\textcolor{black}{commonly known as} North Farm) in Starkville, MS (Figure \ref{figure:3}a). The Husky robot, equipped with \textcolor{black}{all four} cameras, captured images while moving along the cotton rows.

	For \textcolor{black}{perception model} development, images were collected from the OAK-D camera (\textcolor{black}{$1,920  \times  1,080$} pixel resolution), the \textcolor{black}{ZED2i} camera (\textcolor{black}{$1,280  \times  720$} pixel resolution), and the RealSense camera (\textcolor{black}{$640  \times  480$} pixel resolution), as shown in Figures \ref{figure:3}b-d. The dataset included images captured at various angles relative to the horizontal plane ($0 - 45\degree$) and under different weather conditions\textcolor{black}{, such as} cloudy, partially cloudy, and sunny\textcolor{black}{,} to develop a robust \textcolor{black}{vision} model for cotton detection and segmentation.
	
	\begin{table}[ht]
		\caption{Cotton dataset split used for \textcolor{black}{perception} model training, validation, and testing.} 
		\label{table:1a}
		\begin{center}       
			\begin{tabular}{|l|l|l|l|l|l|l|}
				\hline
				\textbf{Dataset} & \textbf{\textcolor{black}{No. of Images}}& \textbf{Split (\%)} & \textbf{No. of Cotton bolls}\\ \hline
				Train & 714 & 70.8\% & 16,852 (70.3\%)\\ \hline
				Validation & 189 & 18.8\% & 4,542 (19.0\%)\\ \hline
				Test & 105 & 10.4\% & 2,566 (10.7\%) \\ \hline
				\textbf{Total} & \textbf{\textcolor{black}{1,008}} &\textbf{100\%} & \textbf{23,960 (100\%)}\\ \hline
			\end{tabular}
		\end{center}
	\end{table}

	Table~\ref{table:1a} presents the final imagery dataset prepared for the initial perception-model training. To ensure diversity and reduce bias, 336 images were randomly selected from each of the three side-facing camera sources, resulting in a total of 1,008 images. These images were selected from the complete dataset, which contained more than 5,000 images, and were combined and randomly mixed before annotation. \textcolor{black}{The selected images were distinct and contained no duplicate or near-duplicate images.} Of the 1,008 images, 714 images (70.8\%) were allocated to the training set, 189 images (18.8\%) were allocated to the validation set, and the remaining 105 images (10.4\%) were allocated to the test set. \textcolor{black}{The three subsets were mutually exclusive, and each image was assigned to only one subset. Therefore, there was no overlap among the training, validation, and test sets.}

	To further evaluate the \textcolor{black}{performance consistency and robustness of the selected models across different data partitions}, five randomized dataset splits were generated using the same pool of 1,008 images, following a 70:10:20 ratio for training, validation, and testing, respectively. Table~\ref{table:1b} presents the average values computed across all five \textcolor{black}{dataset splits, each of which allocated a larger proportion of images to testing than the primary dataset split}. The details of the individual dataset splits are provided in Table~\ref{table:1c}. \textcolor{black}{This evaluation strategy enabled performance averaging across the randomized splits and supported the identification of the model that demonstrated the most consistent performance for subsequent detection and segmentation experiments.}

	\begin{table}[ht]
		\caption{\textcolor{black}{Randomized cotton dataset split structure used to evaluate model performance consistency across five data partitions. Each split allocated 70\% of the images to training, 10\% to validation, and 20\% to testing.}}
		\label{table:1b}
		\begin{center}
			\begin{tabular}{|l|l|l|l|}
				\hline
				\textbf{Dataset} &
				\textbf{\textcolor{black}{No. of Images}} &
				\textbf{Split (\%)} &
				\textbf{\shortstack{\textcolor{black}{A}verage No. of\\
						Cotton \textcolor{black}{B}olls}} \\ \hline
				
				Train & 705 & 70.0\% & 16,771 (70.0\%) \\ \hline
				Validation & 100 & 10.0\% & 2,405 (10.0\%) \\ \hline
				Test & 203 & 20.0\% & 4,784 (20.0\%) \\ \hline
				
				\textbf{Total} &
				\textbf{\textcolor{black}{1,008}} &
				\textbf{100.0\%} &
				\textbf{23,960 (100.0\%)} \\ \hline
			\end{tabular}
		\end{center}
	\end{table}
	
	In addition, a segmentation dataset was created using the `auto\_annotate' module from the Ultralytics library~\cite{ultralytics_auto_annotate}. For this process, the \textcolor{black}{YOLO model that performed best in our preliminary study} was used. This vision model was developed using a smaller and less complex dataset consisting of 300 images. \textcolor{black}{The preliminary dataset containing 300 images was completely independent of the dataset containing 1,008 images used in the present study. The two datasets contained no overlapping images, duplicate images, or nearly identical images.} The choice was restricted to the YOLOv8 and YOLOv11 architectures because the `auto\_annotate' module supports only YOLO models with default pretrained weights available within the Ultralytics framework. \textcolor{black}{Other YOLO variants could not be used because of compatibility limitations with their original repositories.}
	
	Based on its superior detection performance in \textcolor{black}{our preliminary} \textcolor{black}{study}, the YOLOv8-l model was selected for automated annotation \textcolor{black}{for segmentation task}. This model was combined with the SAMv2.1-Hiera-L (Large) pretrained segmentation model to convert the same 1,008-image dataset into a corresponding pixel-wise segmentation dataset. To maintain consistency across all experiments, the same training, validation, and testing data splits described in Table~\ref{table:1a} were applied during segmentation model \textcolor{black}{development}.

	To evaluate the robustness of the top segmentation models from the overall analysis, five randomized dataset splits were also generated using the same 1,008-image dataset, mirroring the structure used for detection (Table~\ref{table:1b}). This ensured that the models could be assessed under consistent conditions, allowing for a fair comparison. \textcolor{black}{The randomized dataset split strategy enabled performance averaging across multiple runs and provided additional evidence of internal performance consistency across different data partitions.}
	
	\subsection{Data Preprocessing}\label{sec:datapreprocessing}
	
\begin{figure}[hbt!]
		\centering
		\begin{subfigure}[b]{0.49\textwidth}
			\centering
			\includegraphics[height=2.5 cm]{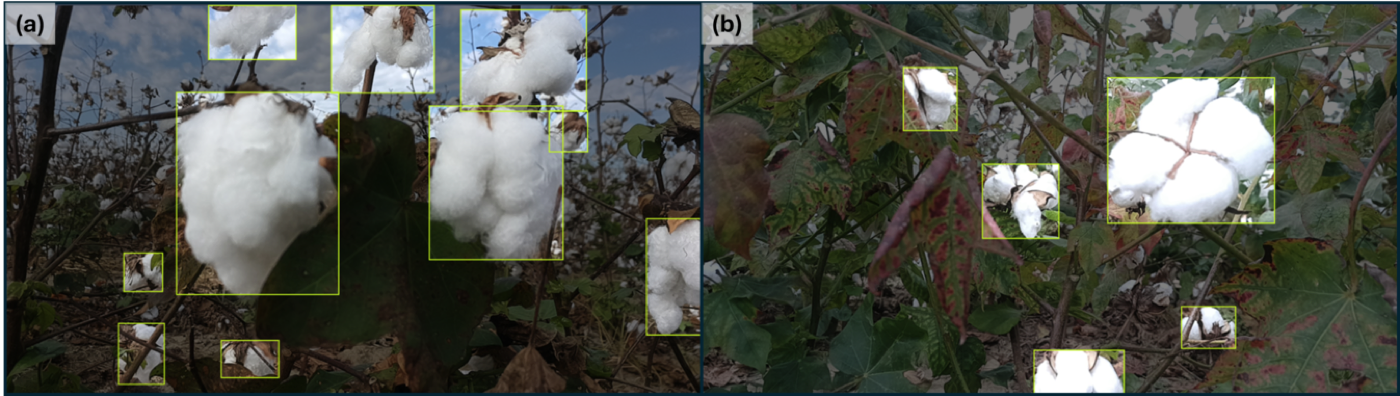}
		\end{subfigure}
		\caption{\textcolor{black}{The visualization of} cotton boll manual \textcolor{black}{annotations} in Roboflow \textcolor{black}{for detection task}}
		\label{figure:4}    
	\end{figure}

\begin{figure}[hbt!]
		\centering
		\begin{subfigure}[b]{0.49\textwidth}
			\centering
			\includegraphics[height=2.5 cm]{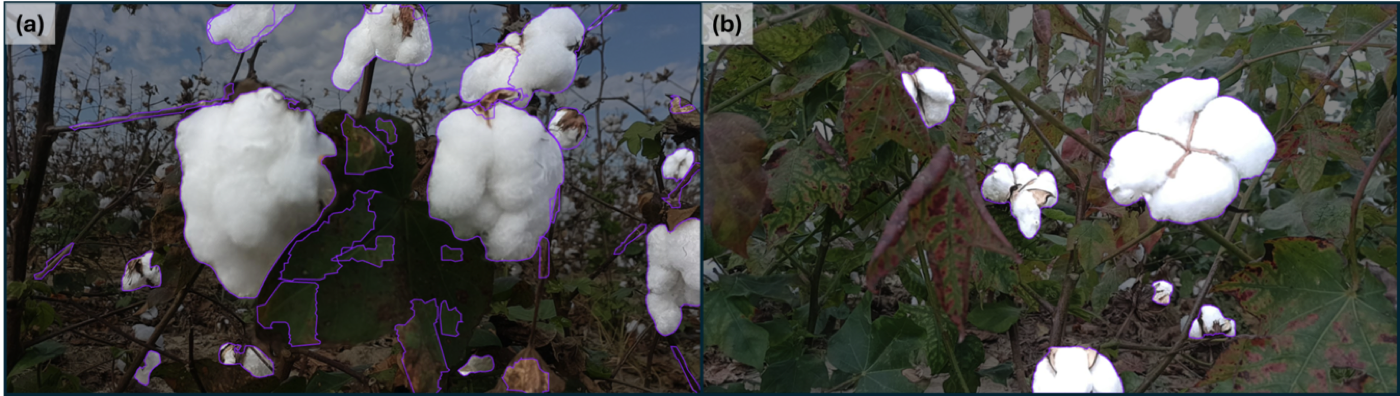}
		\end{subfigure}
		\caption{Visualization of cotton boll segmentation results on Roboflow after the inference.}
		\label{figure:5}    
	\end{figure}

	\textcolor{black}{All images from} detection dataset were manually annotated using Roboflow with a predefined class labeled as `cotton'. This study aims to improve cotton boll detection and segmentation in support of an autonomous cotton-picking system. To align with the robotic system’s physical constraints, only front-row cotton bolls \textcolor{black}{that are} closer to the camera\textcolor{black}{s} and within the manipulator’s reachable workspace were annotated. \textcolor{black}{Our main} focus was on medium to large, visually prominent bolls that are most relevant for \textcolor{black}{robotic} picking. Figure \ref{figure:4} illustrates the annotation strategy and selection criteria applied in Roboflow. Representative images from the resulting segmentation dataset, visualized via the Roboflow platform, are shown in \textcolor{black}{Figure \ref{figure:5}}.

	\subsection{Overall Approach \textcolor{black}{for Segmentation}} \label{sec:overall}

    Two approaches were used to segment the seed cotton: (1) detection followed by segmentation, and (2) direct segmentation. \textcolor{black}{For the detection-prompted segmentation approach, the initial phase focused on identifying the best-performing YOLO model for cotton boll detection using the default model settings. The selected YOLO model and its trained weights were then applied to the test images to generate bounding-box prompts for subsequent segmentation. Candidate detections were retained for segmentation according to their bounding-box dimensions and confidence scores. The Cotton-Eye perception system applied two primary constraints, and a detected cotton boll was retained only when both criteria were satisfied:}
	
	\begin{enumerate}
		\item The cotton boll detection confidence level was required to be greater than or equal to 80\%.
		
		\item Either the width or the height of the cotton boll bounding box was required to be greater than 100 pixels ($w > 100$ pixels or $h > 100$ pixels).
	\end{enumerate}
	
	\textcolor{black}{Therefore, the two primary constraints were combined using an AND condition, whereas the width and height criteria within the second constraint were evaluated using an OR condition. Once the bounding box coordinates satisfying these constraints were extracted, they were passed to the SAM (Segment Anything Model) as prompts to generate segmentation masks. These masks were then used to identify the detected cotton bolls and delineate the corresponding seed cotton regions within the image.} \textcolor{black}{Hereafter, this detection-prompted segmentation pipeline is referred to as the GELAN-s + SAM approach, where GELAN-s provides the detected cotton boll bounding-box prompts and SAM generates the corresponding seed cotton segmentation masks.}
	
	\textcolor{black}{For} the second approach, YOLO segmentation models were trained directly using the derived dataset and applied to segment the seed cotton. \textcolor{black}{The same confidence and bounding box size constraints described above were applied to the direct segmentation approach to retain candidate cotton bolls for subsequent analysis.}

	\subsection{Deep Learning Models}\label{sec:deeplearningmodels}
	
	Computer vision has \textcolor{black}{been advanced significantly} due to the development of Artificial Intelligence (AI) techniques in recent years. Specifically, the development of deep learning models has replaced manual feature extraction \textcolor{black}{process} and provided more reliable outputs. Agriculture is one of the key areas where deep learning models can \textcolor{black}{support the automation of} nearly all farm operations, including autonomous weeding, planting, harvesting, disease detection, postharvest analysis, yield predictions, \textcolor{black}{irrigation scheduling}, and precision farming. Object detection and segmentation \textcolor{black}{models} are continuously evolving and are being applied across \textcolor{black}{different} farming fields to \textcolor{black}{facilitate automation in agriculture.}
	
	\subsubsection{\textcolor{black}{Cotton} Detection Task}

\textcolor{black}{Object detection is a computer vision technique that identifies and localizes objects in images, videos, and live streams using bounding boxes, class labels, and confidence scores. Deep learning-based detectors learn relevant image features from labeled training data to perform these tasks automatically. They are generally categorized as one-stage detectors \cite{wang2022yolov7,liu2016ssd,lin2017focal} or two-stage detectors \cite{girshick2015fast,he2017mask}. YOLO is a widely adopted one-stage detector because of its lightweight architecture and near-real-time detection capabilities. These characteristics make YOLO suitable for agricultural applications requiring rapid and accurate perception.}

\textcolor{black}{In this study, recent variants of the YOLO family, including YOLOv8 \cite{yolov8_ultralytics}, YOLOv9 \cite{wang2024yolov9}, YOLOv10 \cite{wang2024yolov10}, YOLOv11 \cite{yolo11_ultralytics}, YOLOv12 \cite{tian2025yolov12}, and YOLOv13 \cite{lei2025yolov13}, were evaluated for cotton boll detection based on accuracy and inference speed.}

\paragraph{YOLOv8}

\textcolor{black}{YOLOv8 \cite{yolov8_ultralytics} was released in January 2023 with improvements to its backbone and neck architectures for enhanced feature extraction. It also incorporates an anchor-free Ultralytics split head, an optimized accuracy-speed trade-off, and pretrained models of different sizes \cite{docYOLOv8}. Five configurations—YOLOv8-n, YOLOv8-s, YOLOv8-m, YOLOv8-l, and YOLOv8-x—are available with corresponding pretrained weights \cite{yolov8_ultralytics}. All five configurations were evaluated for cotton boll detection using their default hyperparameters and settings.}

\paragraph{YOLOv9}

\textcolor{black}{YOLOv9 improves detection performance through Programmable Gradient Information (PGI) and the Generalized Efficient Layer Aggregation Network (GELAN) \cite{wang2024yolov9}. PGI uses input information to provide reliable gradients for updating network weights during training. GELAN employs gradient-path planning and conventional convolution to support efficient parameter utilization in lightweight models. The combination of PGI and GELAN is intended to balance detection accuracy and computational efficiency. YOLOv9-s, YOLOv9-m, YOLOv9-c, YOLOv9-e, GELAN-s, GELAN-m, GELAN-c, and GELAN-E were evaluated using their default hyperparameters and configurations.}

\paragraph{YOLOv10}

\textcolor{black}{YOLOv10 was released in 2024 with improvements to both post-processing and model architecture \cite{wang2024yolov10}. It uses consistent dual assignment to enable Non-Maximum Suppression (NMS)-free training and low-latency inference. Its efficiency-oriented components include a lightweight classification head, spatial-channel decoupled downsampling, and rank-guided block design. Large-kernel convolution and partial self-attention are incorporated to improve detection accuracy. YOLOv10-n, YOLOv10-s, YOLOv10-m, YOLOv10-b, YOLOv10-l, and YOLOv10-x were evaluated using their default parameters.}

\paragraph{YOLOv11}

\textcolor{black}{YOLOv11 introduces several architectural modifications intended to improve feature extraction and computational efficiency \cite{yolo11_ultralytics}. Its principal components include the Cross Stage Partial with kernel size 2 (C3k2) block, Spatial Pyramid Pooling-Fast (SPPF), and Convolutional block with Parallel Spatial Attention (C2PSA) \cite{khanam2024yolov11}. The C3k2 block replaces the CSP Bottleneck with Two Convolutions, Faster (C2f) block and uses smaller kernel sizes to support faster processing. The C2PSA block following SPPF enhances spatial attention to relevant image regions. YOLOv11-n, YOLOv11-s, YOLOv11-m, YOLOv11-l, and YOLOv11-x were evaluated for cotton boll detection using their default parameters.}

\paragraph{YOLOv12}

\textcolor{black}{YOLOv12 incorporates attention-based components to improve detection accuracy while maintaining high inference speed \cite{tian2025yolov12}. Its two principal components are the Efficient Area Attention (EAA) module and the Recursive Efficient Layer Aggregation Network (R-ELAN). EAA captures information across a larger receptive field while limiting computational overhead. R-ELAN improves learning efficiency and addresses training challenges in deep convolutional networks. Five configurations—YOLOv12-n, YOLOv12-s, YOLOv12-m, YOLOv12-l, and YOLOv12-x—were evaluated using their default hyperparameters.}

\paragraph{YOLOv13}

\textcolor{black}{YOLOv13 \cite{lei2025yolov13} was among the most recent YOLO variants available at the time of this study. It introduces Hypergraph-based Adaptive Correlation Enhancement (HyperACE) to capture high-order correlations and improve cross-location and cross-scale feature fusion. HyperACE was designed to address the limited modeling of global high-order correlations in preceding architectures. YOLOv13 also employs depthwise separable convolutions and lightweight blocks to reduce the number of parameters and floating-point operations (FLOPs). Four configurations—YOLOv13-n, YOLOv13-s, YOLOv13-l, and YOLOv13-x—were evaluated using their default parameters.}

\paragraph{\textcolor{black}{YOLOv26}}

\textcolor{black}{YOLOv26 was not included because it was released in 2026, whereas the comparative analysis and model selection for cotton boll detection were completed in 2025 \cite{jocher2026ultralyticsyolo26unifiedrealtime}. The selected model had subsequently been used in field experiments involving robotic cotton picking.}

\textcolor{black}{The evaluated models were compared as a practical out-of-the-box benchmark using the default configurations provided by their respective implementations. Selected parameters were standardized as described below, whereas implementation-specific default settings were otherwise retained.} The initial learning rate (ILR), learning rate factor (LRF), and box-loss gain were set to their default values of 0.01, 0.01, and 7.5, respectively. Furthermore, the input image size and batch size were set to $1,024 \times 1,024$ pixels and 8, respectively. Early stopping was configured according to the default parameters of the respective model implementations. The YOLOv8 models used an early-stopping patience of 50 epochs, whereas the other models used a patience of 100 epochs. All models were trained for a maximum of 300 epochs. The total number of completed training epochs and the corresponding best-performing epochs are presented in Table~\ref{table:2}. \textcolor{black}{Accordingly, the results should be interpreted as a comparison of the evaluated models under their available default configurations rather than as an isolated comparison of architectural differences under a completely identical training protocol.}

\subsubsection{\textcolor{black}{Cotton} \textcolor{black}{Segmentation} Task}

\textcolor{black}{Image segmentation is a computer vision technique that processes images at the pixel level by assigning each pixel to a specific class or object instance. Deep learning-based segmentation models learn image patterns and generate masks representing object shapes and pixel-level boundaries. The three principal segmentation tasks are semantic, instance, and panoptic segmentation. Semantic segmentation assigns a class label to each pixel, whereas instance segmentation separately identifies and segments individual objects within the same class \cite{chen2018masklab}. Panoptic segmentation combines semantic and instance segmentation within a unified framework \cite{kirillov2019panoptic}.}

In this study, multiple segmentation models were evaluated for their ability to localize cotton bolls through pixel-level mask generation under field conditions. These include promptable segmentation models such as the Segment Anything Model (SAM)~\cite{kirillov2023segment}, SAMv2.1~\cite{ravi2024sam}, FastSAM~\cite{zhao2023fast}, and Grounded-SAM~\cite{ren2024grounded}. In addition, the segmentation capabilities of integrated YOLO-based models, including YOLOv8-seg, YOLOv11-seg, and YOLOv12-seg, were also assessed for cotton boll segmentation performance.

\paragraph{SAM}

\textcolor{black}{The Segment Anything Model (SAM) \cite{kirillov2023segment}, developed by Meta AI Research, supports zero-shot transfer to different segmentation tasks through prompt-based interaction. The model accepts point, bounding-box, and mask prompts and includes an ambiguity-aware design. Segmentation masks are generated by a lightweight mask decoder that combines image embeddings from an image encoder with prompts processed by a prompt encoder. These components allow SAM to produce masks for objects specified by different types of visual prompts. In this study, the SAM ViT-H model \cite{kirillov2023segmentSoft} was used to segment cotton bolls using bounding boxes generated by the best-performing YOLO detector as prompts.}

\paragraph{SAM2}

\textcolor{black}{SAM2 \cite{ravi2024sam} extends the original SAM framework to support both image and video segmentation. It is also reported to be six times faster than SAM for image segmentation. In the SAM2 framework, an image is treated as a single frame within a video sequence, supporting broader generalization through the Promptable Visual Segmentation task. The four SAM2.1 checkpoints evaluated in this study were Tiny, Small, Base Plus, and Large. Segmentation was guided by bounding boxes generated by the best-performing YOLO detector.}

\paragraph{FastSAM}

\textcolor{black}{FastSAM is a real-time alternative to SAM developed to improve segmentation inference speed \cite{zhao2023fast}. It separates the segmentation process into two stages: a CNN-based detector first generates candidate masks, and the second stage filters the masks according to the supplied prompt. FastSAM is built on the YOLOv8-seg architecture, which combines object detection with an instance-segmentation branch. It was trained using 2\% of the SA-1B dataset while achieving segmentation quality comparable to SAM \cite{kirillov2023segment,zhao2023fast}. In this study, pretrained FastSAM-s and FastSAM-x weights were evaluated, while SAM was guided by bounding boxes generated by the best-performing YOLO detector. The text prompts ``cotton'' and ``cotton bolls'' were used for FastSAM, and the inference script retained the 20 masks with the highest confidence scores from each image.}

\paragraph{Grounded-SAM}

\textcolor{black}{Grounded-SAM combines Grounding DINO with the SAM architecture to provide both open-vocabulary detection and segmentation capabilities \cite{ren2024grounded,kirillov2023segment,liu2023grounding}. Its integration with the Recognize Anything Model (RAM) can support automatic image-annotation pipelines \cite{huang2023inject,zhang2023recognize,huang2023tag2text}. This combination allows relevant objects to be recognized, localized, and segmented using complementary vision models. In this study, Grounded-SAM was combined with RAM and evaluated for seed-cotton segmentation and its potential integration into robotic cotton-picking systems.}

\paragraph{YOLO Segmentation}

\textcolor{black}{Recent YOLO models incorporate instance-segmentation capabilities in addition to object detection, extending their use in near-real-time applications. Although YOLOv8, YOLOv11, and YOLOv12 were primarily developed for object detection, their segmentation variants include dedicated heads for pixel-level mask generation. YOLOv8-seg has previously outperformed Mask R-CNN \cite{he2017mask} in precision, recall, and inference speed in complex orchard environments \cite{sapkota2024comparing}. In this study, the Ultralytics implementations of YOLOv8-seg, YOLOv11-seg, and YOLOv12-seg were trained on the segmentation dataset and compared to identify the most effective model for seed-cotton segmentation.}

	\subsection{Evaluation, \textcolor{black}{V}alidation, and \textcolor{black}{M}etrics}\label{sec:evaluaiton}

	\textcolor{black}{The detection models were evaluated using bounding box precision, recall, F1 score, and mean average precision at an Intersection over Union threshold of 0.5 (mAP@0.5). The instance segmentation models were evaluated using the corresponding mask metrics. These metrics followed standard object detection and instance segmentation evaluation procedures~\cite{everingham2010pascal,lin2014microsoft}.}
	
	\textcolor{black}{This study considered one object class, namely open cotton boll. Therefore, all reported metrics represent the performance of the models in detecting or segmenting open cotton bolls.}
	
	\begin{equation}
		\label{eq:1}
		Precision = \frac{TP}{TP+FP}
	\end{equation}
	
	\begin{equation}
		\label{eq:2}
		Recall = \frac{TP}{TP+FN}
	\end{equation}
	
	\begin{equation}
		\label{eq:3}
		\textcolor{black}{
			F1 =
			\frac{2 \times Precision \times Recall}
			{Precision + Recall}
		}
	\end{equation}
	
	\textcolor{black}{Intersection over Union (IoU) measures the overlap between a predicted region and its corresponding ground truth region and is calculated as follows:}
	
	\begin{equation}
		\label{eq:iou}
		\textcolor{black}{
			IoU =
			\frac{\mathrm{Area}(R_{\mathrm{pred}} \cap R_{\mathrm{gt}})}
			{\mathrm{Area}(R_{\mathrm{pred}} \cup R_{\mathrm{gt}})}
		}
	\end{equation}
	
	\textcolor{black}{Here, $R_{\mathrm{pred}}$ and $R_{\mathrm{gt}}$ represent the predicted and ground truth regions, respectively. For the detection models, these regions were bounding boxes. For the instance segmentation models, these regions were segmentation masks. At an IoU threshold $\tau$, a prediction was classified as a True Positive (TP) when it had the correct class and achieved an IoU greater than or equal to $\tau$ with an unmatched ground truth instance. An unmatched prediction was classified as a False Positive (FP), while an unmatched ground truth instance was classified as a False Negative (FN). One to one matching was used so that each ground truth instance could be matched with only one prediction.}
	
	\textcolor{black}{For a specified IoU threshold $\tau$, predictions for class $c$ were ranked in descending order according to their confidence scores. Precision and recall were then calculated as the confidence threshold varied. The interpolated precision at recall level $r$ was defined as the maximum precision obtained at any recall level greater than or equal to $r$:}
	
	\begin{equation}
		\label{eq:interp_precision}
		\textcolor{black}{
			p_{\mathrm{interp},c}(r,\tau)
			=
			\max_{\tilde{r}\geq r}p_c(\tilde{r},\tau)
		}
	\end{equation}
	
	\textcolor{black}{The Average Precision (AP) for class $c$ at IoU threshold $\tau$ was defined as the area under the interpolated precision and recall curve~\cite{everingham2010pascal,lin2014microsoft}:}
	
	\begin{equation}
		\label{eq:4}
		\textcolor{black}{
			AP_c@\tau
			=
			\int_{0}^{1}p_{\mathrm{interp},c}(r,\tau)\,dr
		}
	\end{equation}
	
	\noindent\textcolor{black}{where $p_c(r,\tau)$ denotes the precision for class $c$ at recall level $r$ and IoU threshold $\tau$.}
	
	\textcolor{black}{The integral was numerically approximated using the interpolation procedure implemented by the corresponding model evaluation framework. Mean Average Precision at a fixed IoU threshold was calculated by averaging the AP values across the evaluated classes:}
	
	\begin{equation}
		\label{eq:4a}
		\textcolor{black}{
			mAP@\tau
			=
			\frac{1}{N_c}
			\sum_{c=1}^{N_c}AP_c@\tau
		}
	\end{equation}
	
	\noindent\textcolor{black}{where $N_c$ denotes the number of evaluated object classes.}
	
	\textcolor{black}{Because the present study contained only one class, the reported mAP at a fixed IoU threshold was numerically equivalent to the AP of the open cotton boll class:}
	
	\begin{equation}
		\label{eq:single_class_map}
		\textcolor{black}{
			mAP@\tau
			=
			AP_{\mathrm{open\ cotton\ boll}}@\tau,
			\qquad N_c=1
		}
	\end{equation}
	
	\textcolor{black}{An IoU threshold of 0.5 was used for all reported mAP@0.5 values. Detection mAP@0.5 was based on bounding box IoU, whereas segmentation mAP@0.5 was based on mask IoU.}

	\textcolor{black}{Moreover, the agreement between the model predictions and the reference values was evaluated using the coefficient of determination ($R^2$), as presented in Equations~\ref{eq:6} and~\ref{eq:7}. An $R^2$ value closer to 1 indicates better agreement. A negative $R^2$ value indicates that the prediction errors exceed those obtained when the mean reference value is used as the baseline prediction.}

	\begin{equation} \label{eq:6}
		R^2 = 1 - \frac{SSE}{SST}
	\end{equation}
	
	\begin{equation} \label{eq:7}
		R^2 = 1 - 
		\frac{\textcolor{black}{\sum_{i=1}^{N}(y_i-\hat{y}_i)^2}}
		{\textcolor{black}{\sum_{i=1}^{N}(y_i-\bar{y})^2}}
	\end{equation}
	
	\noindent\textcolor{black}{where $N$ represents the total number of evaluated images, $y_i$ represents the reference value for the $i$th image, $\hat{y}_i$ represents the corresponding predicted value, and $\bar{y}$ represents the mean of the reference values. For the count analysis, these variables represent cotton boll counts. For the area analysis, they represent total segmented areas. In Equation~\ref{eq:6}, $SSE$ represents the sum of squared errors, while $SST$ represents the total sum of squares of the reference values.}
	
	\textcolor{black}{Furthermore}, to complement the mAP evaluation and support the selection of the best-performing models, Frames Per Second (FPS) was also analyzed to assess each model’s real-time performance. FPS is a critical metric that quantifies how many frames a model can process per second, offering insight into its suitability for real-time deployment. Equation \ref{eq:8} \textcolor{black}{calculates} FPS based on the total number of processed frames and the corresponding inference time\textcolor{black}{, where a higher value suggests better inference speed.}
	
	\begin{equation} \label{eq:8}
		FPS = \frac{\mathrm{\textcolor{black}{Total}\ number\ of\ frames}}
		{\mathrm{\textcolor{black}{Total}\ processing\ time\ in\ seconds}}
	\end{equation}

	A combined mAP vs. FPS plot was also generated to visualize the relationship between detection accuracy and inference speed. This analysis provides a comprehensive understanding of each model's balance between accuracy and real time deployment suitability.
	
	\textcolor{black}{Before generating the normalized mAP@0.5 vs. FPS plot and calculating the combined performance score, the mAP@0.5 and FPS values were independently rescaled to a range between 0 and 1 using min--max normalization based on the values observed across the evaluated models. This transformation placed the two metrics on comparable scales and prevented the larger numerical magnitude of FPS from disproportionately influencing the combined score. Because min--max normalization is a linear and monotonic transformation, it preserved the model ordering and relative separation within each metric. The normalization was calculated as follows:}
	
	\begin{equation}
		\label{eq:9}
		\textcolor{black}{
			x_{i,\mathrm{norm}}
			=
			\frac{x_i-x_{\min}}
			{x_{\max}-x_{\min}}
		}
	\end{equation}
	
	\noindent\textcolor{black}{where $x_i$ represents the original mAP@0.5 or FPS value for model $i$, $x_{\min}$ represents the minimum value of the corresponding metric, and $x_{\max}$ represents its maximum value.}
	
	\textcolor{black}{A combined score was then calculated by adding the normalized mAP@0.5 and normalized FPS values:}
	
	\begin{equation}
		\label{eq:10}
		\textcolor{black}{
			S_i
			=
			\mathrm{mAP}_{i,\mathrm{norm}}
			+
			\mathrm{FPS}_{i,\mathrm{norm}}
		}
	\end{equation}
	
	\noindent\textcolor{black}{where $S_i$ represents the combined performance score for model $i$. Models with both normalized mAP@0.5 and normalized FPS values greater than or equal to 0.45 were retained as candidates, ensuring that neither metric fell below the specified relative screening threshold. This threshold was used only to compare models within the evaluated model set and was not interpreted as an absolute measure of model quality or deployment suitability. The combined score was subsequently used to rank the candidate models for the randomized dataset split evaluation.}
	
	In addition, \textcolor{black}{cotton} segmentation models were evaluated using both standard segmentation metrics and spatial area comparisons against a manually annotated ground-truth test dataset to assess their effectiveness in seed cotton segmentation. Quantifying segmentation masks using area is inherently challenging, particularly in scenes with overlapping or closely clustered objects. To ensure reliable ground-truth measurements, the 105 test images were manually \textcolor{black}{and carefully} annotated using precise polygonal masks in Roboflow. The total segmented area produced by each model\textcolor{black}{,} including the auto-annotated test dataset\textcolor{black}{,} was compared against these manual annotations. The proposed area-calculation approach explicitly avoided double counting by considering only unique, non-overlapping mask regions. In addition, regression analysis was performed to assess the correspondence between predicted and ground-truth segmentation areas, providing further insight into \textcolor{black}{cotton} segmentation accuracy and consistency.
	
	Furthermore, the top-performing model identified from the evaluation was \textcolor{black}{further} deployed \textcolor{black}{as a part of the Cotton-Eye perception pipeline} during the initial field tests for cotton \textcolor{black}{localization} and picking at the \textcolor{black}{University of Georgia's} Hort Hill Fields in Tifton, Georgia, conducted \textcolor{black}{during} November 10--14, 2025. \textcolor{black}{The YOLOv12-m-seg model, selected based on the accuracy--speed trade-off analysis, was used for the field-level evaluation. The field-testing system consisted of the developed cotton picker equipped with a UR5e robotic arm, the developed cotton-picking end-effector, and a ZED2i stereo camera for cotton boll perception and localization.}

	\textcolor{black}{During the field tests, all field-level inference experiments were conducted on a workstation equipped with an AMD Ryzen 7 5000-series processor (8 cores, 16 threads; Zen 3). The computing system includes an NVIDIA GeForce RTX 3070 Laptop GPU with 8~GB of dedicated GDDR6 memory, providing CUDA and Tensor Core acceleration for deep learning inference. System memory consists of 64~GB of DDR4 RAM operating at 3200 million data transfers per second.}

	The perception evaluation during field testing included assessing the number of detected cotton bolls at confidence levels of 80\% and 70\%, as well as recording the total numbers of accepted and rejected detections. The 80\% confidence level was used as a stricter detection criterion, whereas the 70\% confidence level was included to retain more detections and \textcolor{black}{document system operation under a less restrictive confidence threshold.}
	
	\textcolor{black}{For each confidence threshold, a detected cotton boll was accepted as an actionable robotic-picking target only when all three operational criteria were satisfied: (1) the detection confidence met or exceeded the selected threshold, (2) either the width or height of the bounding box was greater than 100 pixels, and (3) the estimated distance of the cotton boll from the camera was no greater than 1.00~m. A detection was rejected when one or more of these criteria were not satisfied, including cases in which the detected cotton boll was located outside the accessible workspace of the robotic manipulator.}

	A Dell Precision 7920 Tower workstation (Dell Inc., Round Rock, TX, USA), equipped with an Intel$^{\circledR}$ Xeon Gold 6250 CPU (up to 3.90~GHz), 187~GB of system memory, and an NVIDIA$^{\circledR}$ RTX A6000 GPU with 48~GB of dedicated GDDR6 memory, was used to perform all \textcolor{black}{offline} training\textcolor{black}{,} validation, and \textcolor{black}{test-set evaluation} tasks in this study. The workstation was running Ubuntu 20.04.5 LTS.
	
\section{RESULTS}

\textcolor{black}{The results are presented in four stages: detection-model screening and selection, evaluation of direct YOLO-based segmentation, comparison of direct and detection-prompted segmentation approaches, and field evaluation of the selected perception model.}

\subsection{\textcolor{black}{Cotton Boll Detection Model Screening and Selection}}

\textcolor{black}{Table~\ref{table:2} summarizes the training and validation performance of the evaluated cotton boll detection models. YOLOv9-s achieved the highest mAP@0.5 and precision, with values of 84.8\% and 81.0\%, respectively. YOLOv12-s achieved the highest recall of 77.1\%, while YOLOv9-s and GELAN-s shared the highest F1 score of 77.9\%. The complete model-specific performance and training results are reported in Table~\ref{table:2}.}

\begin{table*}[ht]
	\caption{Training and validation results of the YOLO models for \textcolor{black}{c}otton boll detection. IoU refers to \textcolor{black}{Intersection over Union}. \textcolor{black}{The maximum number of training epochs was set to 300.}}
	\label{table:2}
	\begin{center}
		\begin{tabular}{|l|l|l|l|l|l|}
			\hline
			\textbf{YOLO Model} &
			\textbf{\textcolor{black}{mAP@0.5 (\%)}} &
			\textbf{Precision (\%)} &
			\textbf{Recall (\%)} &
			\textbf{\textcolor{black}{F1 Score (\%)}} &
			\textbf{Epochs (Best \textcolor{black}{Weight})}\\
			\hline
			YOLOv8-n & 81.0 & 78.7 & 72.3 & 75.4 & 117 (67)\\ \hline
			YOLOv8-s & 73.3 & 75.0 & 67.0 & 70.8 & 80 (80)\\ \hline
			YOLOv8-m & 82.3 & 77.6 & 74.3 & 75.9 & 86 (36)\\ \hline
			YOLOv8-l & 73.1 & 77.9 & 65.2 & 71.0 & 88 (38)\\ \hline
			YOLOv8-x & 74.6 & 77.6 & 67.0 & 71.9 & 99 (49)\\ \hline
			\textbf{YOLOv9-s} & \textbf{84.8} & \textbf{81.0} & 75.0 & \textbf{77.9} & 275 (175)\\ \hline
			YOLOv9-m & 83.3 & 79.1 & 74.3 & 76.6 & 148 (48)\\ \hline
			YOLOv9-c & 83.5 & 79.9 & 74.7 & 77.2 & 141 (41)\\ \hline
			YOLOv9-e & 81.1 & 79.2 & 75.9 & 77.5 & 145 (45)\\ \hline
			\textbf{GELAN-s} & 84.7 & 80.3 & 75.7 & \textbf{77.9} & 234 (134)\\ \hline
			GELAN-m & 83.4 & 78.4 & 75.7 & 77.0 & 144 (44)\\ \hline
			GELAN-c & 82.7 & 77.9 & 74.5 & 76.2 & 138 (38)\\ \hline
			GELAN-e & 81.7 & 79.5 & 74.5 & 76.9 & 299 (298)\\ \hline
			YOLOv10-n & 79.7 & 77.5 & 71.1 & 74.2 & 177 (77)\\ \hline
			YOLOv10-s & 81.3 & 77.2 & 73.7 & 75.4 & 158 (58)\\ \hline
			YOLOv10-m & 81.1 & 78.5 & 73.3 & 75.8 & 175 (75)\\ \hline
			YOLOv10-b & 81.4 & 76.6 & 73.6 & 75.1 & 155 (55)\\ \hline
			YOLOv10-l & 81.9 & 76.5 & 74.0 & 75.2 & 138 (38)\\ \hline
			YOLOv10-x & 81.0 & 78.2 & 71.8 & 74.9 & 145 (45)\\ \hline
			YOLOv11-n & 82.2 & 79.0 & 72.6 & 75.7 & 159 (59)\\ \hline
			YOLOv11-s & 82.7 & 79.4 & 73.7 & 76.4 & 164 (64)\\ \hline
			YOLOv11-m & 79.6 & 79.0 & 71.6 & 75.1 & 168 (68)\\ \hline
			YOLOv11-l & 80.3 & 77.5 & 71.8 & 74.5 & 151 (51)\\ \hline
			YOLOv11-x & 83.0 & 79.2 & 73.8 & 76.4 & 146 (46)\\ \hline
			YOLOv12-n & 82.0 & 77.0 & 73.1 & 75.0 & 139 (39)\\ \hline
			\textbf{YOLOv12-s} & 83.8 & 77.5 & \textbf{77.1} & 77.3 & 180 (80)\\ \hline
			YOLOv12-m & 83.8 & 79.5 & 74.3 & 76.8 & 148 (48)\\ \hline
			YOLOv12-l & 83.6 & 80.3 & 73.9 & 77.0 & 148 (48)\\ \hline
			YOLOv12-x & 83.1 & 80.2 & 73.7 & 76.8 & 181 (81)\\ \hline
			YOLOv13-n & 81.5 & 77.9 & 72.3 & 75.0 & 126 (26)\\ \hline
			YOLOv13-s & 82.8 & 77.2 & 74.7 & 75.9 & 142 (42)\\ \hline
			YOLOv13-l & 83.5 & 78.4 & 75.0 & 76.7 & 124 (24)\\ \hline
			YOLOv13-x & 82.2 & 77.6 & 73.6 & 75.5 & 141 (41)\\
			\hline
		\end{tabular}
	\end{center}
\end{table*}

\textcolor{black}{The trained detection models were subsequently evaluated using the test dataset, as summarized in Table~\ref{table:3}. YOLOv9-s achieved the highest test mAP@0.5, recall, and F1 score, with values of 86.3\%, 77.6\%, and 79.2\%, respectively. GELAN-e achieved the highest precision of 83.9\%, \textcolor{black}{whereas YOLOv10-m achieved the highest inference speed of approximately 33.0~FPS, corresponding to an inference latency of 30.3~ms per image.} The complete test results for all evaluated models are reported in Table~\ref{table:3}.}

\begin{table*}[ht]
	\caption{Testing results of the YOLO models for \textcolor{black}{c}otton boll detection. IoU refers to \textcolor{black}{Intersection over Union}. \textcolor{black}{Inference speed was calculated using model inference time only and excluded image preprocessing and output postprocessing.}}
	\label{table:3}
	\begin{center}
		\begin{tabular}{|l|l|l|l|l|l|}
			\hline
			\textbf{YOLO Model} &
			\textbf{\textcolor{black}{mAP@0.5 (\%)}} &
			\textbf{Precision (\%)} &
			\textbf{Recall (\%)} &
			\textbf{\textcolor{black}{F1 Score (\%)}} &
			\textbf{\textcolor{black}{Inference Speed per Image (ms)}}\\
			\hline
			YOLOv8-n & 82.2 & 80.5 & 72.3 & 76.2 & 32.3\\ \hline
			YOLOv8-s & 82.5 & 80.0 & 73.8 & 76.8 & 45.0\\ \hline
			YOLOv8-m & 83.7 & 80.9 & 74.2 & 77.4 & 43.7\\ \hline
			YOLOv8-l & 84.3 & 79.9 & 75.5 & 77.6 & 46.1\\ \hline
			YOLOv8-x & 83.6 & 80.2 & 75.3 & 77.7 & 54.2\\ \hline
			\textbf{YOLOv9-s} & \textbf{86.3} & 80.9 & \textbf{77.6} & \textbf{79.2} & 47.8\\ \hline
			YOLOv9-m & 84.2 & 82.5 & 74.4 & 78.2 & 57.8\\ \hline
			YOLOv9-c & 84.1 & 80.3 & 75.2 & 77.7 & 60.8\\ \hline
			YOLOv9-e & 84.1 & 78.0 & 77.2 & 77.6 & 63.6\\ \hline
			GELAN-s & 86.1 & 81.6 & 76.6 & 79.0 & 42.3\\ \hline
			GELAN-m & 84.5 & 79.1 & 76.8 & 77.9 & 50.6\\ \hline
			GELAN-c & 83.6 & 80.6 & 73.2 & 76.7 & 53.8\\ \hline
			\textbf{GELAN-e} & 82.9 & \textbf{83.9} & 73.9 & 78.6 & 64.4\\ \hline
			YOLOv10-n & 81.6 & 79.1 & 72.6 & 75.7 & 36.1\\ \hline
			YOLOv10-s & 82.9 & 79.7 & 74.3 & 76.9 & 37.6\\ \hline
			\textbf{YOLOv10-m} & 82.6 & 80.9 & 73.8 & 77.2 & \textbf{30.3}\\ \hline
			YOLOv10-b & 82.8 & 80.2 & 72.8 & 76.3 & 41.2\\ \hline
			YOLOv10-l & 83.2 & 81.1 & 72.2 & 76.4 & 34.6\\ \hline
			YOLOv10-x & 83.1 & 81.7 & 71.6 & 76.3 & 47.5\\ \hline
			YOLOv11-n & 82.4 & 78.3 & 73.1 & 75.6 & 43.4\\ \hline
			YOLOv11-s & 83.4 & 80.2 & 73.8 & 76.9 & 44.2\\ \hline
			YOLOv11-m & 84.2 & 79.8 & 76.5 & 78.1 & 46.0\\ \hline
			YOLOv11-l & 83.4 & 79.6 & 73.5 & 76.4 & 53.3\\ \hline
			YOLOv11-x & 84.1 & 82.5 & 72.2 & 77.0 & 52.6\\ \hline
			YOLOv12-n & 80.2 & 79.1 & 71.0 & 74.8 & 39.3\\ \hline
			YOLOv12-s & 85.0 & 81.1 & 77.1 & 79.0 & 44.4\\ \hline
			YOLOv12-m & 84.2 & 79.8 & 75.9 & 77.8 & 42.8\\ \hline
			YOLOv12-l & 84.7 & 79.6 & 76.1 & 77.8 & 52.1\\ \hline
			YOLOv12-x & 84.4 & 81.0 & 76.1 & 78.5 & 57.6\\ \hline
			YOLOv13-n & 81.0 & 78.8 & 73.3 & 76.0 & 45.2\\ \hline
			YOLOv13-s & 80.5 & 78.7 & 72.9 & 75.7 & 51.7\\ \hline
			YOLOv13-l & 82.4 & 78.2 & 75.2 & 76.7 & 59.6\\ \hline
			YOLOv13-x & 81.1 & 79.5 & 73.6 & 76.4 & 72.3\\
			\hline
		\end{tabular}
	\end{center}
\end{table*}

\begin{figure}[hbt!]
	\centering
	\begin{subfigure}[b]{0.49\textwidth}
		\centering
		\includegraphics[height=8cm]{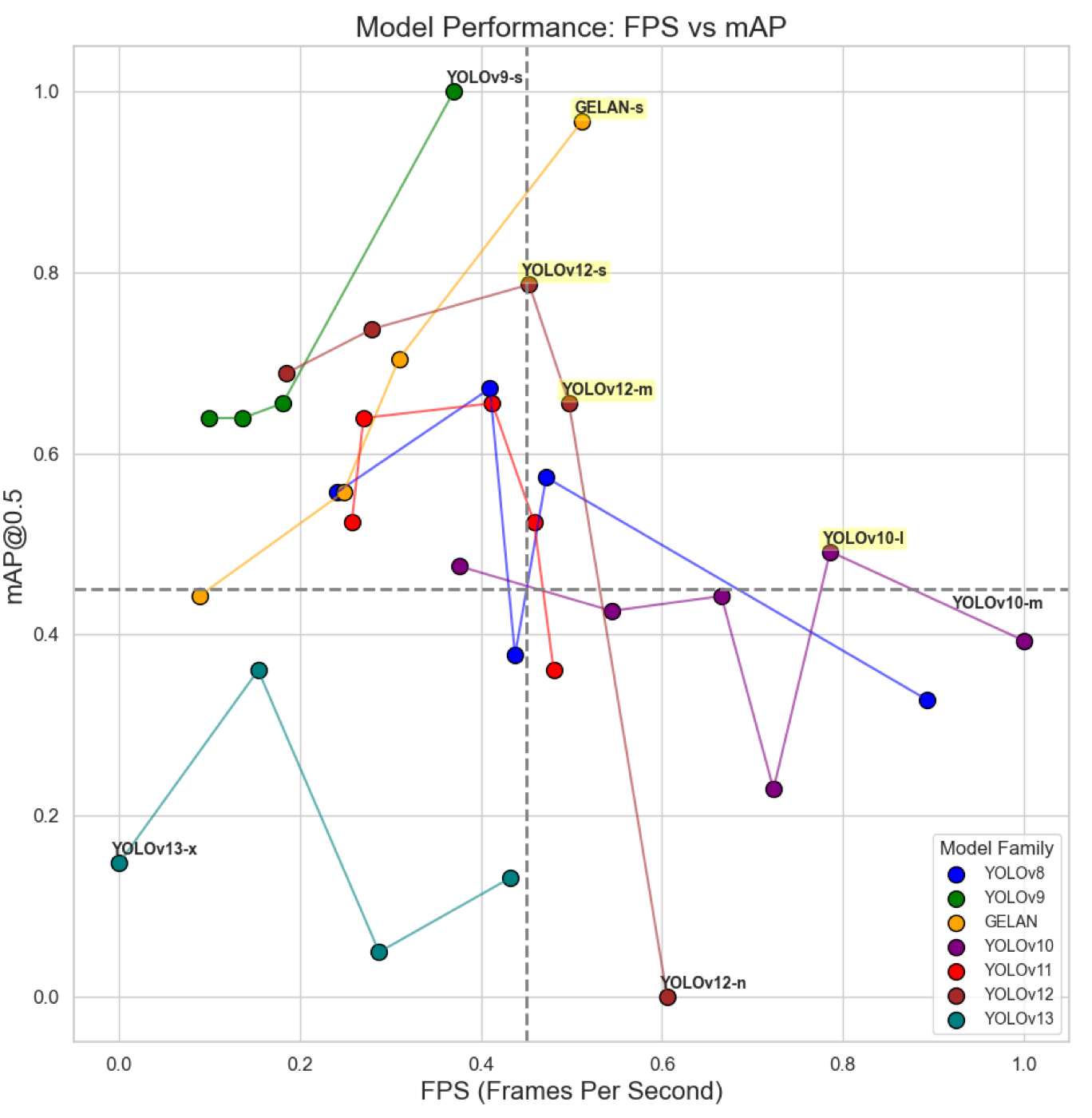}
	\end{subfigure}
	\caption{\textcolor{black}{Normalized mAP@0.5 and FPS values of the cotton boll detection models evaluated using the test dataset.} \textcolor{black}{Each marker group represents a YOLO model family, and each dot within a group represents a distinct detection-model variant. All evaluated variants are displayed, but only the candidate variants and selected lower-performing variants are labeled to prevent label overlap.} \textcolor{black}{The dashed horizontal and vertical lines indicate the reference value of 0.45 for normalized mAP@0.5 and normalized FPS, respectively.} \textcolor{black}{Model variants exceeding both reference values were retained for the randomized dataset split evaluation.}}
	\label{figure:7}
\end{figure}

\textcolor{black}{The detection models were next screened according to their test-set accuracy and inference speed. Figure~\ref{figure:7} compares the normalized mAP@0.5 and FPS values, using a reference value of 0.45 for both metrics. Models exceeding both reference values were retained as candidates and ranked using the combined score defined in Equation~\ref{eq:10}. This procedure selected GELAN-s, YOLOv10-l, YOLOv12-s, and YOLOv12-m for evaluation across the five randomized dataset splits.}

\textcolor{black}{The four candidate models were further evaluated across the five randomized dataset splits described in Tables~\ref{table:1b} and~\ref{table:1c}, which were generated from the original dataset presented in Table~\ref{table:1a}. Tables~\ref{table:6} and~\ref{table:7} report the average training, validation, and testing results, while the complete results for the individual splits are provided in Tables~\ref{table:8} and~\ref{table:9} in the appendix.}

\begin{table*}[ht]
	\caption{\textcolor{black}{Average training and validation results of the four selected cotton boll detection models across five randomized dataset splits. The candidate models were identified using the screening procedure shown in Figure~\ref{figure:7}. Results are presented as the mean $\pm$ sample standard deviation. IoU refers to Intersection over Union. The maximum number of training epochs was set to 300.}}
	\label{table:6}
	\begin{center}       
		
		\begin{tabular}{|l|l|l|l|l|}
			\hline
			\textbf{YOLO Model} & 
			\textbf{mAP (\%) (IoU=0.5)} & 
			\textbf{Precision (\%)} & 
			\textbf{Recall (\%)} & 
			\textbf{F1-Score \textcolor{black}{(\%)}} \\ \hline
			
			\textbf{GELAN-s} & 
			\textbf{85.3 \textcolor{black}{$\pm$ 0.6}} & 
			\textbf{80.8 \textcolor{black}{$\pm$ 1.1}} & 
			75.8 \textcolor{black}{$\pm$ 0.5} & 
			\textbf{78.2 \textcolor{black}{$\pm$ 0.5}} \\ \hline
			
			YOLOv10-l & 
			82.3 \textcolor{black}{$\pm$ 0.4} & 
			78.0 \textcolor{black}{$\pm$ 1.4} & 
			73.9 \textcolor{black}{$\pm$ 1.0} & 
			75.9 \textcolor{black}{$\pm$ 0.4} \\ \hline
			
			YOLOv12-s & 
			83.5 \textcolor{black}{$\pm$ 0.7} & 
			79.5 \textcolor{black}{$\pm$ 1.9} & 
			74.3 \textcolor{black}{$\pm$ 1.7} & 
			76.8 \textcolor{black}{$\pm$ 0.8} \\ \hline 
			
			\textbf{YOLOv12-m} & 
			83.5 \textcolor{black}{$\pm$ 0.6} & 
			77.3 \textcolor{black}{$\pm$ 1.3} & 
			\textbf{76.3 \textcolor{black}{$\pm$ 1.4}} & 
			76.8 \textcolor{black}{$\pm$ 0.5} \\ \hline 		
			
		\end{tabular}
	\end{center}
\end{table*}

\begin{table*}[ht]
	\caption{\textcolor{black}{Average testing results of the four selected cotton boll detection models across five randomized dataset splits. The candidate models were identified using the screening procedure shown in Figure~\ref{figure:7}. Results are presented as the mean $\pm$ sample standard deviation. IoU refers to Intersection over Union. FPS refers to frames per second.}}
	\label{table:7}
	\begin{center}
		
		\resizebox{\textwidth}{!}{%
			\begin{tabular}{|l|l|l|l|l|l|l|}
				\hline
				\textbf{YOLO Model} &
				\textbf{mAP (\%) (IoU=0.5)} &
				\textbf{Precision (\%)} &
				\textbf{Recall (\%)} &
				\textbf{F1-Score \textcolor{black}{(\%)}} &
				\textbf{Inference \textcolor{black}{S}peed \textcolor{black}{P}er \textcolor{black}{I}mage (ms)} &
				\textbf{FPS} \\ \hline
				
				\textbf{GELAN-s} &
				\textbf{84.5 \textcolor{black}{$\pm$ 0.7}} &
				\textbf{79.2 \textcolor{black}{$\pm$ 1.4}} &
				\textbf{76.1 \textcolor{black}{$\pm$ 0.5}} &
				\textbf{77.6 \textcolor{black}{$\pm$ 0.7}} &
				26.7 \textcolor{black}{$\pm$ 2.2} &
				\textcolor{black}{37.68} \textcolor{black}{$\pm$ 3.09} \\ \hline
				
				YOLOv10-l &
				82.2 \textcolor{black}{$\pm$ 0.7} &
				\textcolor{black}{77.2} \textcolor{black}{$\pm$ 0.6} &
				74.5 \textcolor{black}{$\pm$ 1.6} &
				75.6 \textcolor{black}{$\pm$ 0.7} &
				28.8 \textcolor{black}{$\pm$ 3.0} &
				35.02 \textcolor{black}{$\pm$ 3.54} \\ \hline
				
				\textcolor{black}{\textbf{YOLOv12-s}} &
				82.7 \textcolor{black}{$\pm$ 0.9} &
				78.7 \textcolor{black}{$\pm$ 1.5} &
				73.4 \textcolor{black}{$\pm$ 0.8} &
				75.9 \textcolor{black}{$\pm$ 0.9} &
				\textbf{25.0 \textcolor{black}{$\pm$ 3.0}} &
				\textbf{40.54 \textcolor{black}{$\pm$ 5.22}} \\ \hline
				
				YOLOv12-m &
				82.9 \textcolor{black}{$\pm$ 0.6} &
				77.4 \textcolor{black}{$\pm$ 1.2} &
				75.2 \textcolor{black}{$\pm$ 1.0} &
				76.3 \textcolor{black}{$\pm$ 0.8} &
				29.5 \textcolor{black}{$\pm$ 2.3} &
				34.10 \textcolor{black}{$\pm$ 2.59} \\ \hline
				
			\end{tabular}%
		}
	\end{center}
\end{table*}

\textcolor{black}{Across the five randomized splits, GELAN-s achieved the highest average training and validation mAP@0.5, precision, and F1 score, with values of 85.3\%, 80.8\%, and 78.2\%, respectively. YOLOv12-m achieved the highest average recall of 76.3\% (Table~\ref{table:6}). In the test-set evaluation, GELAN-s achieved the highest average mAP@0.5, precision, recall, and F1 score, with values of 84.5\%, 79.2\%, 76.1\%, and 77.6\%, respectively. YOLOv12-s provided the shortest average inference time of 25.0~ms per image, corresponding to 40.54~FPS (Table~\ref{table:7}). Based on these results, GELAN-s was selected to generate the bounding-box prompts used in the subsequent detection-prompted segmentation evaluation. A qualitative comparison of the four candidate detection models under the operational confidence and bounding-box size criteria is provided in Figure~\ref{figure:8} in the Appendix.}

\subsection{\textcolor{black}{Direct YOLO-Based Segmentation Model Screening and Selection}}

\textcolor{black}{Table~\ref{table:4} summarizes the training and validation performance of the direct YOLO-based segmentation models using the inferred masks. YOLOv8-x-seg achieved the highest mAP@0.5 of 84.5\%, YOLOv11-x-seg achieved the highest precision of 85.3\%, YOLOv12-m-seg achieved the highest recall of 77.3\%, and YOLOv11-l-seg achieved the highest F1 score of 80.0\%. The complete training and validation results are reported in Table~\ref{table:4}.}

\begin{table*}[ht]
	\caption{Training and validation results of the YOLO segmentation models for \textcolor{black}{c}otton boll \textcolor{black}{segmentation}. IoU refers to \textcolor{black}{Intersection over Union}. \textcolor{black}{The maximum number of training epochs was set to 300.}}
	\label{table:4}
	\begin{center}
		\begin{tabular}{|l|l|l|l|l|l|}
			\hline
			\textbf{YOLO Model} &
			\textbf{\textcolor{black}{mAP@0.5 (\%)}} &
			\textbf{Precision (\%)} &
			\textbf{Recall (\%)} &
			\textbf{\textcolor{black}{F1 Score (\%)}} &
			\textbf{Epochs (Best \textcolor{black}{Weight})}\\
			\hline
			YOLOv8-n-seg & 82.0 & 80.5 & 73.7 & 77.0 & 138 (88)\\ \hline
			YOLOv8-s-seg & 82.9 & 80.1 & 75.2 & 77.6 & 92 (42)\\ \hline
			YOLOv8-m-seg & 83.6 & 81.1 & 75.4 & 78.2 & 174 (124)\\ \hline
			YOLOv8-l-seg & 84.3 & 84.7 & 75.2 & 79.7 & 269 (219)\\ \hline
			\textbf{YOLOv8-x-seg} & \textbf{84.5} & 82.7 & 75.9 & 79.2 & 95 (45)\\ \hline
			YOLOv11-n-seg & 82.6 & 81.7 & 74.2 & 77.7 & 300 (300)\\ \hline
			YOLOv11-s-seg & 83.5 & 83.4 & 75.2 & 79.1 & 300 (300)\\ \hline
			YOLOv11-m-seg & 83.5 & 84.0 & 74.2 & 78.8 & 300 (300)\\ \hline
			\textbf{YOLOv11-l-seg} & 84.4 & 84.4 & 76.1 & \textbf{80.0} & 300 (300)\\ \hline
			\textbf{YOLOv11-x-seg} & 84.0 & \textbf{85.3} & 74.6 & 79.6 & 300 (300)\\ \hline
			YOLOv12-n-seg & 82.5 & 80.8 & 74.5 & 77.5 & 224 (124)\\ \hline
			YOLOv12-s-seg & 82.8 & 82.0 & 75.0 & 78.3 & 300 (300)\\ \hline
			\textbf{YOLOv12-m-seg} & 84.1 & 81.3 & \textbf{77.3} & 79.2 & 300 (300)\\ \hline
			YOLOv12-l-seg & 84.1 & 82.0 & 77.1 & 79.5 & 300 (300)\\ \hline
			YOLOv12-x-seg & 83.7 & 81.5 & 76.9 & 79.1 & 300 (300)\\
			\hline
		\end{tabular}
	\end{center}
\end{table*}

\textcolor{black}{The direct YOLO-based segmentation models were subsequently evaluated using the test set with inferred segmentation masks (Table~\ref{table:5}). YOLOv8-l-seg achieved the highest test mAP@0.5 and F1 score, with values of 84.5\% and 80.1\%, respectively. YOLOv11-s-seg achieved the highest precision of 84.4\%, while YOLOv11-x-seg achieved the highest recall of 77.3\%. YOLOv12-n-seg provided the shortest inference time of 14.8~ms per image. The complete test results are reported in Table~\ref{table:5}.}

\begin{table*}[ht]
	\caption{Testing results of the YOLO segmentation models for \textcolor{black}{c}otton boll \textcolor{black}{segmentation}. IoU refers to \textcolor{black}{Intersection over Union}. \textcolor{black}{Inference speed was calculated using model inference time only and excluded image preprocessing and output postprocessing.}}
	\label{table:5}
	\begin{center}
		\begin{tabular}{|l|l|l|l|l|l|}
			\hline
			\textbf{YOLO Model} &
			\textbf{\textcolor{black}{mAP@0.5 (\%)}} &
			\textbf{Precision (\%)} &
			\textbf{Recall (\%)} &
			\textbf{\textcolor{black}{F1 Score (\%)}} &
			\textbf{\textcolor{black}{Inference Speed per Image (ms)}}\\
			\hline
			YOLOv8-n-seg & 82.7 & 82.8 & 74.3 & 78.3 & 24.5\\ \hline
			YOLOv8-s-seg & 81.8 & 81.3 & 73.8 & 77.4 & 23.0\\ \hline
			YOLOv8-m-seg & 84.1 & 84.0 & 75.3 & 79.4 & 26.0\\ \hline
			\textbf{YOLOv8-l-seg} & \textbf{84.5} & 84.1 & 76.4 & \textbf{80.1} & 28.4\\ \hline
			YOLOv8-x-seg & 84.0 & 82.7 & 75.3 & 78.8 & 33.4\\ \hline
			YOLOv11-n-seg & 83.3 & 83.0 & 74.5 & 78.5 & 25.6\\ \hline
			\textbf{YOLOv11-s-seg} & 82.4 & \textbf{84.4} & 73.9 & 78.8 & 44.2\\ \hline
			YOLOv11-m-seg & 83.9 & 82.6 & 76.2 & 79.3 & 26.7\\ \hline
			YOLOv11-l-seg & 83.9 & 82.8 & 76.1 & 79.3 & 32.3\\ \hline
			\textbf{YOLOv11-x-seg} & 83.9 & 81.7 & \textbf{77.3} & 79.4 & 34.6\\ \hline
			\textbf{\textcolor{black}{YOLOv12-n-seg}} & 82.8 & 81.7 & 76.0 & 78.7 & \textbf{\textcolor{black}{14.8}}\\ \hline
			YOLOv12-s-seg & 82.8 & 83.5 & 75.1 & 79.1 & 15.7\\ \hline
			YOLOv12-m-seg & 83.7 & 81.3 & 77.1 & 79.1 & 20.4\\ \hline
			YOLOv12-l-seg & 83.7 & 81.4 & 77.2 & 79.2 & 24.7\\ \hline
			YOLOv12-x-seg & 84.0 & 82.3 & 76.8 & 79.5 & 34.0\\
			\hline
		\end{tabular}
	\end{center}
\end{table*}

\begin{figure}[hbt!]
	\centering
	\begin{subfigure}[b]{0.49\textwidth}
		\centering
		\includegraphics[height=8cm]{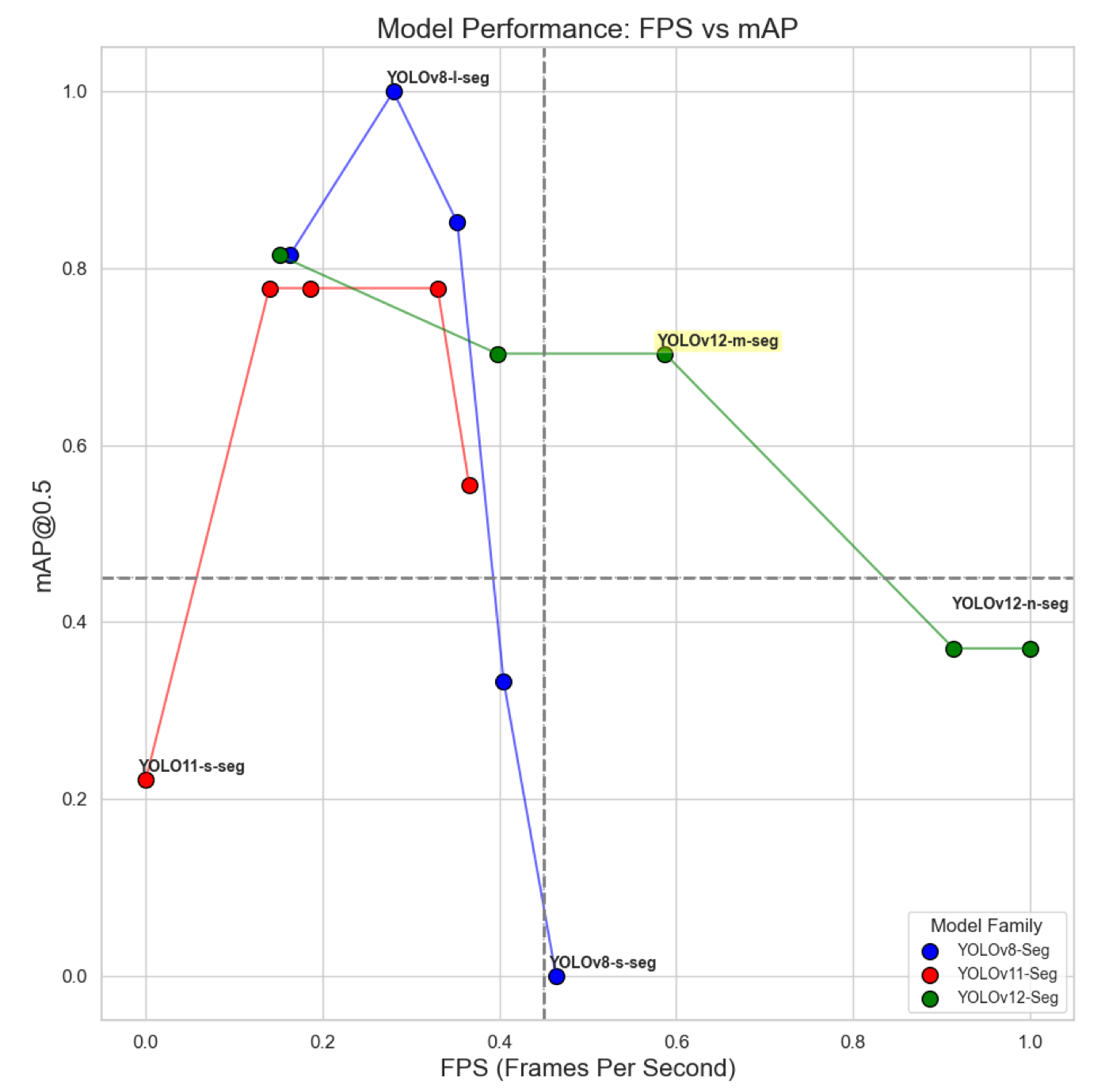}
	\end{subfigure}
	\caption{\textcolor{black}{Normalized mAP@0.5 and FPS values of the direct YOLO-based cotton boll segmentation models.} \textcolor{black}{Each marker group represents a YOLO model family, and each dot within a group represents a distinct segmentation-model variant. All evaluated variants are displayed, but only the candidate variants and selected lower-performing variants are labeled to prevent label overlap.} \textcolor{black}{The dashed horizontal and vertical lines indicate the reference value of 0.45 for normalized mAP@0.5 and normalized FPS, respectively.} \textcolor{black}{YOLOv12-m-seg was the only model variant that exceeded both reference values.}}
	\label{figure:9}
\end{figure}

\textcolor{black}{The direct segmentation models were next screened according to their test-set segmentation accuracy and inference speed. Figure~\ref{figure:9} compares the normalized mAP@0.5 and FPS values using a reference value of 0.45 for both metrics. YOLOv12-m-seg was the only model that exceeded both reference values and was therefore retained for evaluation across the randomized dataset splits and for subsequent comparison with the detection-prompted segmentation approach.}

\textcolor{black}{YOLOv12-m-seg was subsequently evaluated across the same five randomized image subsets used for the detection-model evaluation. The average training and validation mAP@0.5, precision, recall, and F1 score were 84.2\%, 81.0\%, 77.2\%, and 79.1\%, respectively (Table~\ref{table:10}). Across the corresponding test sets, the average mAP@0.5, precision, recall, and F1 score were 84.0\%, 81.6\%, 76.2\%, and 78.8\%, respectively (Table~\ref{table:11}). Representative YOLOv12-m-seg outputs under sunny, partially cloudy, and cloudy field conditions are provided in Figure~\ref{figure:10} in the Appendix. In the representative sunny example, part of one large cotton boll was not segmented, whereas all visible large cotton bolls were segmented in the representative partially cloudy and cloudy examples. These observations apply only to the illustrated images.}

\subsection{\textcolor{black}{Detection-Prompted Segmentation Screening and Selection}}

\textcolor{black}{The selected GELAN-s detection model was combined with SAM using two segmentation strategies: direct bounding-box prompting and segmentation of clipped image patches generated from the detected regions. A representative comparison of these strategies is provided in Figure~\ref{figure:11} in the Appendix. In the clipped-patch approach, portions of cotton leaves and background regions, including clouds, were incorporated into the segmentation masks. Based on this qualitative comparison, direct bounding-box prompting was retained for the subsequent SAM-based evaluations.}

\textcolor{black}{The Tiny, Small, Base+, and Large SAMv2.1 variants~\cite{ravi2024sam} were then evaluated using bounding-box prompts generated by GELAN-s for representative sunny, partially cloudy, and cloudy field images. Their qualitative outputs are provided in Figure~\ref{figure:13} in the Appendix, and their total segmented mask areas are reported in Table~\ref{table:areanalysis}. The segmented areas generally increased from the Tiny variant to the larger variants, although the pattern was not monotonic across the three images. Because no statistical test was conducted, these differences are reported only as a descriptive comparison.}

\begin{table}[ht]
	\caption{\textcolor{black}{Total cotton segmentation mask areas produced by the SAMv2.1 variants using GELAN-s detection-based bounding-box prompts for representative sunny, partially cloudy, and cloudy field images.}}
	\label{table:areanalysis}
	\centering
	\begin{tabular}{|>{\raggedright\arraybackslash}p{2cm}|l|l|}
		\hline
		\makecell[l]{\textbf{\textcolor{black}{Field Weather}}\\\textbf{Condition}} &
		\textbf{SAMv2.1 Variant} &
		\makecell{\textbf{\textcolor{black}{Total} Mask Area}\\\textbf{(pixels)}} \\ \hline
		
		\multirow{4}{*}{\makecell[l]{Sunny\\(Figure~\ref{figure:13}a)}} 
		& Tiny      & 547,393 \\ \cline{2-3}
		& Small     & 554,781 \\ \cline{2-3}
		& Base+ & 562,734 \\ \cline{2-3}
		& Large     & 559,820 \\ \hline
		
		\multirow{4}{*}{\makecell[l]{Partially cloudy\\(Figure~\ref{figure:13}b)}} 
		& Tiny      & 452,420 \\ \cline{2-3}
		& Small     & 458,813 \\ \cline{2-3}
		& Base+ & 452,527 \\ \cline{2-3}
		& Large     & 459,190 \\ \hline
		
		\multirow{4}{*}{\makecell[l]{Cloudy\\(Figure~\ref{figure:13}c)}} 
		& Tiny      & 444,678 \\ \cline{2-3}
		& Small     & 434,572 \\ \cline{2-3}
		& Base+ & 446,498 \\ \cline{2-3}
		& Large     & 451,522 \\ \hline
	\end{tabular}
\end{table}

\textcolor{black}{The qualitative masks were largely similar among the four variants, and the differences in total segmented area were comparatively small. SAMv2.1 Tiny was therefore retained as the lightweight SAMv2.1 variant for the subsequent model comparisons.}

\subsection{\textcolor{black}{Comparison and Selection of the Final Segmentation Approach}}

\begin{figure}[hbt!]
	\centering
	\begin{subfigure}[b]{0.50\textwidth}
		\centering
		\includegraphics[height=5cm]{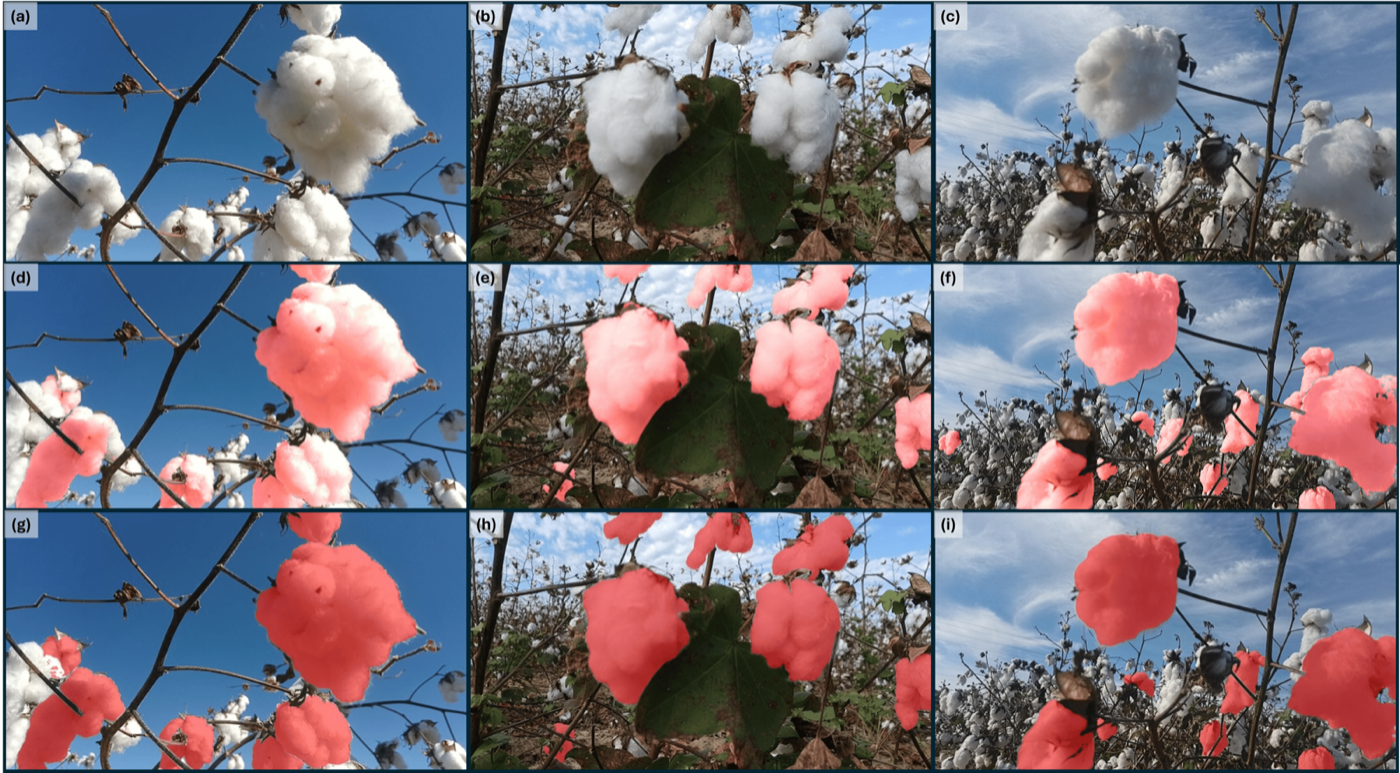}
	\end{subfigure}
	\caption{\textcolor{black}{Qualitative comparison of cotton boll segmentation produced by GELAN-s + SAMv2.1 Tiny and YOLOv12-m-seg. Panels (a), (b), and (c) show representative sunny, partially cloudy, and cloudy field images, respectively. Panels (d)--(f) show the corresponding GELAN-s + SAMv2.1 Tiny results, while panels (g)--(i) show the YOLOv12-m-seg results.}}
	\label{figure:15}
\end{figure}

\textcolor{black}{Figure~\ref{figure:15} presents a qualitative comparison of the two retained segmentation approaches: GELAN-s + SAMv2.1 Tiny and YOLOv12-m-seg. In the representative sunny image, the two approaches produced comparable segmentation outputs. In the partially cloudy and cloudy examples, both approaches segmented the front-row cotton bolls that met the confidence and bounding-box size criteria. GELAN-s + SAMv2.1 Tiny also segmented several back-row cotton bolls in these two examples. This observation indicates greater sensitivity to additional cotton bolls in the illustrated scenes but does not establish greater robustness across the complete dataset.}

\begin{figure*}[ht]
	\centering
	\begin{subfigure}[b]{0.49\textwidth}
		\centering
		\includegraphics[width=\linewidth]{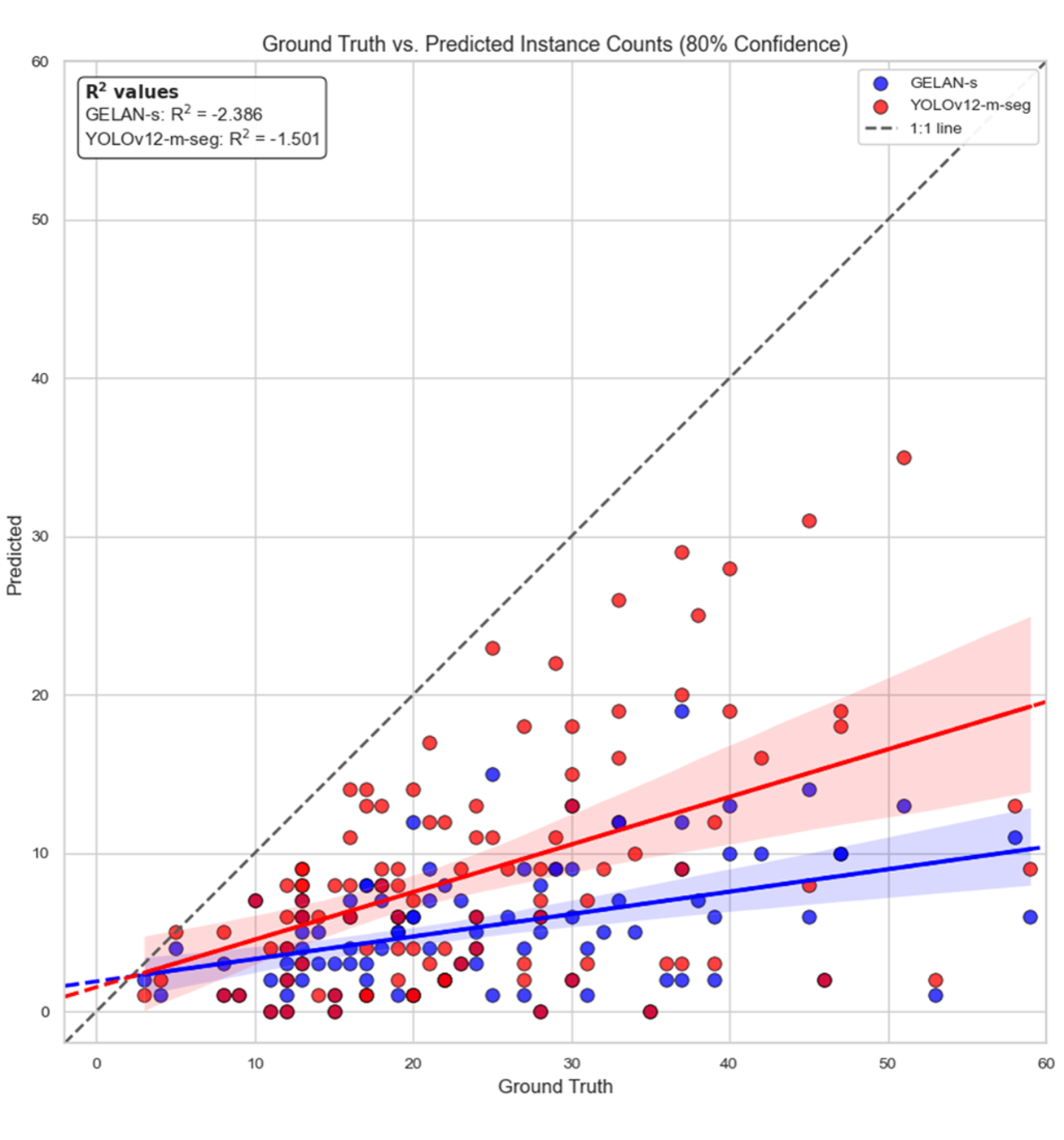}
		\caption{\textcolor{black}{Confidence level of 80\%.}}
		\label{figure:16a}
	\end{subfigure}
	\hfill
	\begin{subfigure}[b]{0.49\textwidth}
		\centering
		\includegraphics[width=\linewidth]{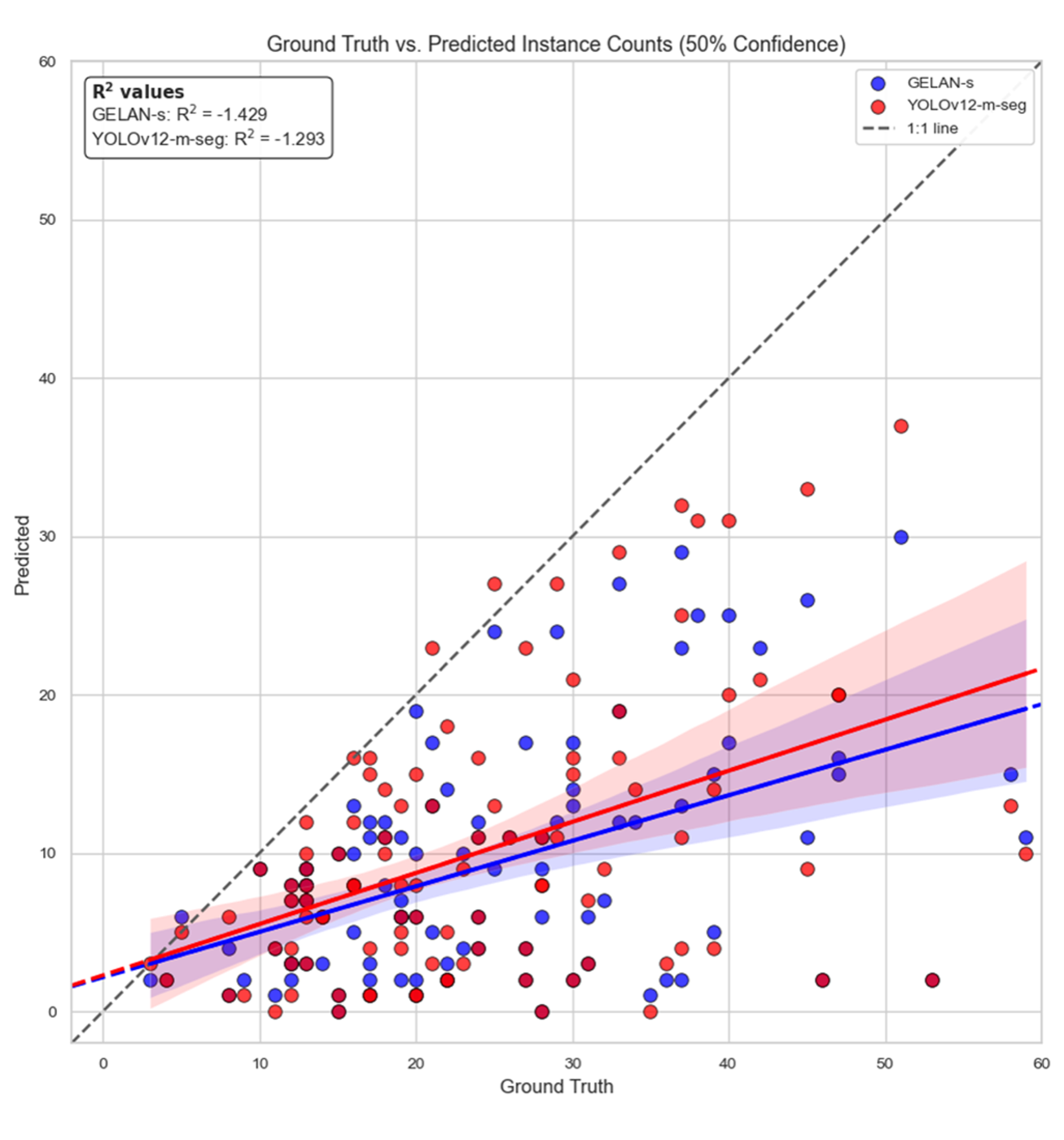}
		\caption{\textcolor{black}{Confidence level of 50\%.}}
		\label{figure:16b}
	\end{subfigure}
	\caption{\textcolor{black}{Comparison of cotton boll counts produced by GELAN-s and YOLOv12-m-seg with the reference counts derived from the automatically annotated test masks at confidence levels of (a) 80\% and (b) 50\%.}}
	\label{figure:16}
\end{figure*}

\textcolor{black}{The two retained approaches were quantitatively compared using the complete test dataset of 105 images. Cotton boll counts produced by GELAN-s and YOLOv12-m-seg were compared with the reference counts derived from the automatically annotated test masks at confidence levels of 80\% and 50\% (Figure~\ref{figure:16}). At the 80\% confidence level, GELAN-s and YOLOv12-m-seg produced $R^2$ values of $-2.386$ and $-1.501$, respectively. At the 50\% confidence level, their corresponding $R^2$ values were $-1.429$ and $-1.293$. All four values were negative, indicating poor absolute agreement with the reference counts. Although YOLOv12-m-seg produced less-negative values at both confidence levels, these results do not establish reliable cotton boll counting.}

\begin{figure}[hbt!]
	\centering
	\begin{subfigure}[b]{0.50\textwidth}
		\centering
		\includegraphics[height=8cm]{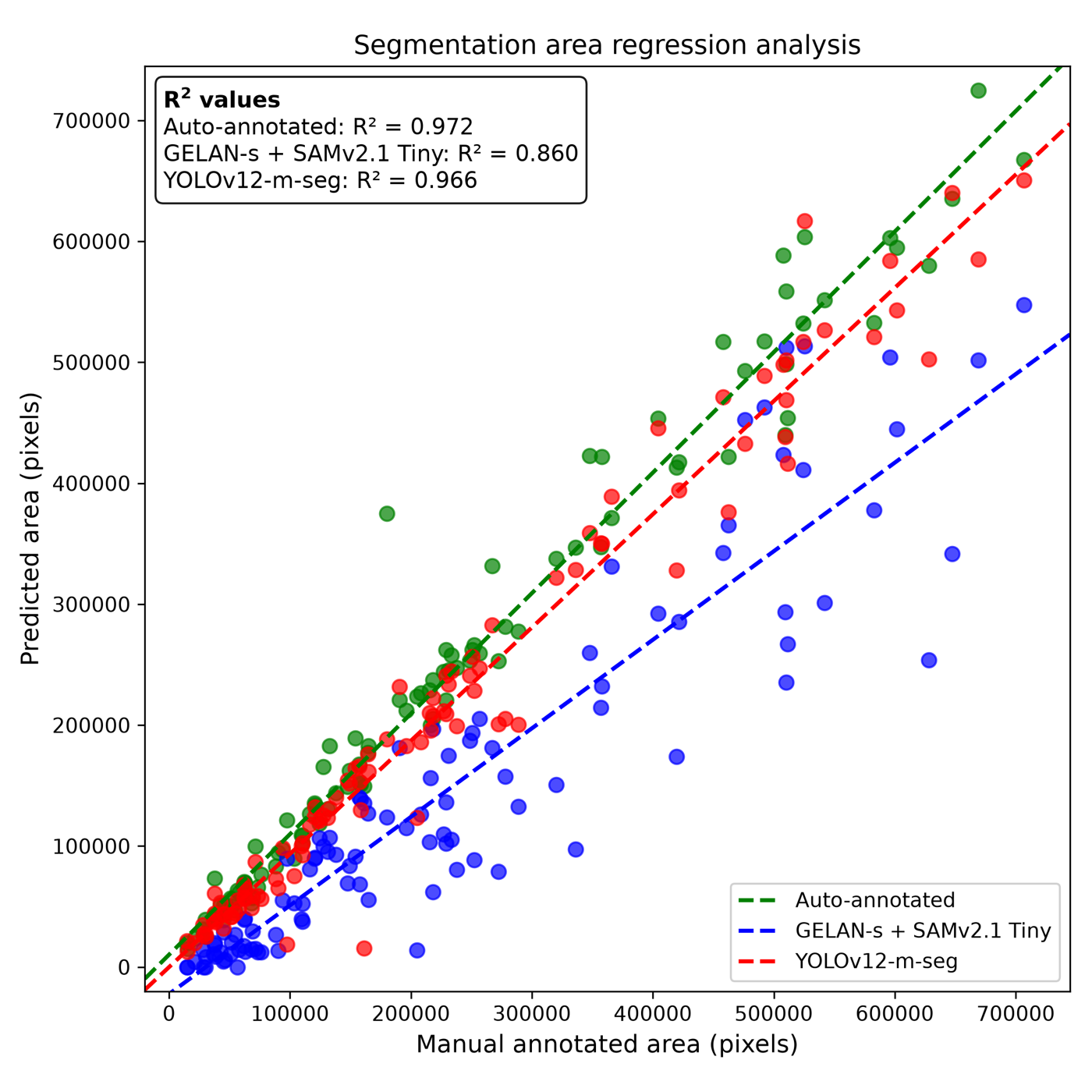}
	\end{subfigure}
	\caption{\textcolor{black}{Agreement between the manually annotated total cotton segmentation area and the areas obtained from the automatically annotated masks, YOLOv12-m-seg, and GELAN-s + SAMv2.1 Tiny. Segmented areas are reported in pixels.}}
	\label{figure:17}
\end{figure}

\textcolor{black}{Segmented-area agreement was evaluated separately using manually annotated masks as the reference (Figure~\ref{figure:17}). The automatically annotated masks achieved an $R^2$ value of 0.972, YOLOv12-m-seg achieved 0.966, and GELAN-s + SAMv2.1 Tiny achieved 0.860. The GELAN-s + SAMv2.1 Tiny approach also generally underestimated the total segmented cotton area.}

\textcolor{black}{The inference-speed comparison showed that YOLOv12-m-seg required 20.4~ms per image, whereas GELAN-s + SAMv2.1 Tiny required approximately 60~ms per image. The YOLOv12-m-seg inference time is also reported in Table~\ref{table:5}. Based on its less-negative count-based $R^2$ values, greater segmented-area agreement, and shorter inference time, YOLOv12-m-seg was selected for field evaluation.}

\subsection{\textcolor{black}{Field Evaluation of the Selected Segmentation Model}}

\textcolor{black}{YOLOv12-m-seg was deployed with the robotic cotton-picking system for field evaluation. A detection was accepted when it met the selected confidence threshold, had a bounding-box width or height greater than 100 pixels, and was located within 1.00~m of the camera. Detections that did not meet the operational criteria or were outside the robot workspace were rejected.}

\textcolor{black}{The field outcomes are summarized in Table~\ref{table:fieldtest}.} \textcolor{black}{At the 80\% confidence level, 797 of the 1,627 detected cotton bolls (49.0\%) were accepted as actionable picking targets, whereas 830 (51.0\%) were rejected. At the 70\% confidence level, 981 of the 1,706 detected cotton bolls (57.5\%) were accepted and 725 (42.5\%) were rejected. The available field-test records retained only the final acceptance or rejection status and did not identify the specific criterion responsible for each rejection. Consequently, the rejected detections could not be retrospectively disaggregated according to size, distance, or manipulator-workspace constraints. Future field evaluations will incorporate rule-specific rejection logging to identify the primary factors limiting target acceptance.}

\begin{table}[ht]
	\caption{\textcolor{black}{Field outcomes of the cotton-picking system implemented with YOLOv12-m-seg. Cotton boll detections are reported as accepted or rejected according to the operational confidence, bounding-box size, distance, and robot-workspace criteria.}}
	\label{table:fieldtest}
	\begin{center}
		\begin{tabular}{|l|l|l|l|}
			\hline
			\makecell{\textbf{Confidence}\\\textbf{Level}} &
			\textbf{Date \textcolor{black}{of Field Test}} &
			\textbf{Accepted} &
			\textbf{Rejected}\\
			\hline
			\multirow{4}{*}{80\%} & 2025-11-11 & 47 & 148\\ \cline{2-4}
			& 2025-11-12 & 66 & 230\\ \cline{2-4}
			& 2025-11-13 & 684 & 452\\ \cline{2-4}
			& \textbf{Total} & \textbf{797} & \textbf{830}\\
			\hline
			\multirow{3}{*}{70\%} & 2025-11-13 & 430 & 302\\ \cline{2-4}
			& 2025-11-14 & 551 & 423\\ \cline{2-4}
			& \textbf{Total} & \textbf{981} & \textbf{725}\\
			\hline
		\end{tabular}
	\end{center}
\end{table}

\begin{figure}[hbt!]
	\centering
	\begin{subfigure}[b]{0.50\textwidth}
		\centering
		\includegraphics[height=6.5cm]{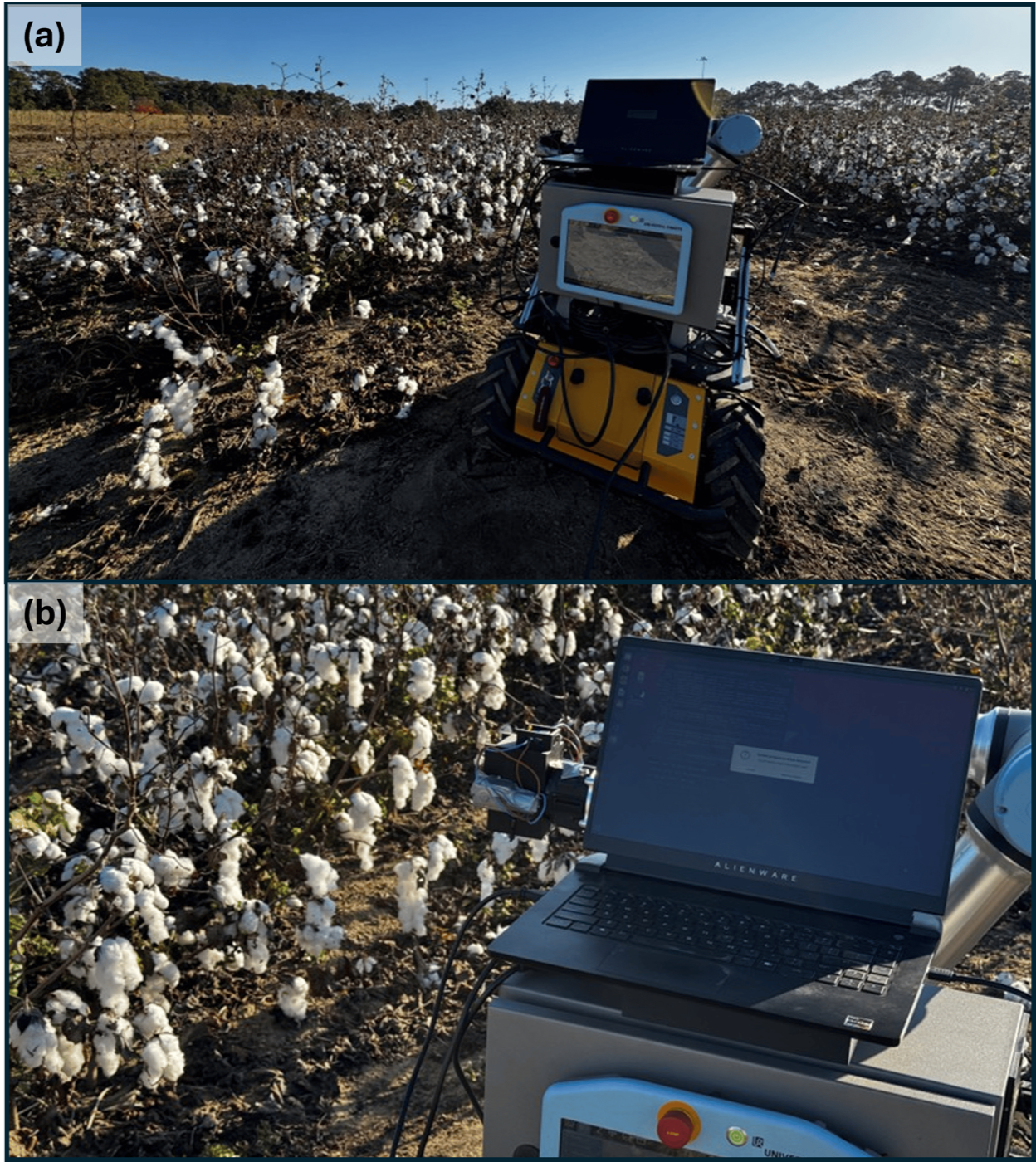}
	\end{subfigure}
	\caption{\textcolor{black}{Field evaluation of the developed robotic cotton picker using the Cotton-Eye perception system implemented with YOLOv12-m-seg. The experiments were conducted at the University of Georgia Hort Hill fields in Tifton, Georgia, from November 10 to 14, 2025.}}
	\label{figure:18}
\end{figure}

\begin{figure}[hbt!]
	\centering
	\begin{subfigure}[b]{0.49\textwidth}
		\centering
		\includegraphics[height=6.5cm]{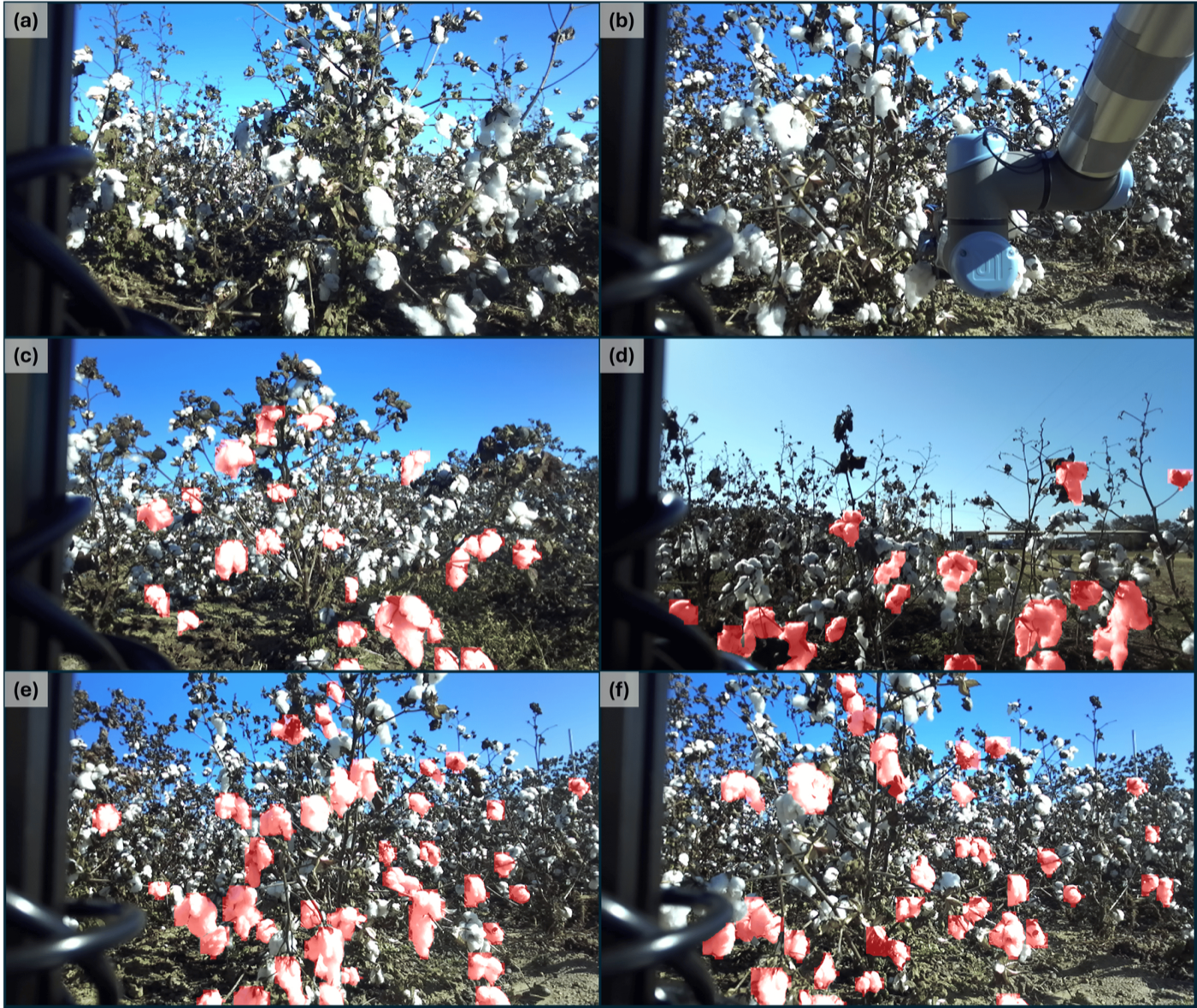}
	\end{subfigure}
	\caption{\textcolor{black}{Field deployment of the Cotton-Eye perception system using YOLOv12-m-seg. Panel (a) shows a representative field image captured using the ZED2i stereo camera, and panel (b) shows the UR5e manipulator performing cotton picking using the perception-system outputs. Panels (c)--(d) show segmentation outputs recorded during trials at an 80\% confidence level, while panels (e)--(f) show outputs recorded during trials at a 70\% confidence level.}}
	\label{figure:19}
\end{figure}

\textcolor{black}{The overall field-evaluation setup is shown in Figure~\ref{figure:18}. Figure~\ref{figure:19} shows a representative image captured with the ZED2i stereo camera, the UR5e manipulator performing cotton picking using the localized target coordinates, and YOLOv12-m-seg outputs recorded during field trials at confidence levels of 80\% and 70\%.}

\section{DISCUSSION}

\textcolor{black}{The discussion focuses on three major outcomes of this study. First, the cotton boll detection models are discussed in terms of accuracy, consistency across the randomized dataset splits, and inference speed. Second, the direct YOLO-based segmentation and GELAN-s-guided SAM approaches are compared. Finally, the selected YOLOv12-m-seg model is discussed based on its deployment with the robotic cotton-picking system.}

\textcolor{black}{Relative to previous cotton studies that commonly focused on individual components, including boll detection and counting, image segmentation, yield estimation, or specific stages of robotic harvesting~\cite{liu2023small,zhang2024yolo,reddy2024cotton,gharakhani2024field}, the distinguishing contribution of this work is the integration of model benchmarking, evaluation of accuracy and inference speed, and field implementation within a unified perception framework. Specifically, this study benchmarks recent YOLO detection models, direct YOLO segmentation models, and SAM approaches guided by detection results. It evaluates model accuracy, consistency across randomized dataset splits, and inference speed. The selected YOLOv12-m-seg model is then deployed within the Cotton-Eye perception pipeline under field conditions.}

\subsection{\textcolor{black}{Cotton Boll Detection Model Performance}}

\textcolor{black}{YOLOv9-s achieved the highest mAP@0.5, recall, and F1 score on the initial test dataset, while GELAN-e achieved the highest precision (Table~\ref{table:3}). However, the evaluation across five randomized dataset splits showed that GELAN-s achieved the highest average mAP@0.5, precision, recall, and F1 score (Table~\ref{table:7}). YOLOv12-s provided the shortest average inference time among the four candidate models. Based on the combined evaluation of the accuracy--speed trade-off and performance across the randomized datasets, GELAN-s was selected to generate the bounding-box prompts for the subsequent SAM-based segmentation evaluation.}

\textcolor{black}{The differences between the current results and the initial unpublished experiments may be related to the increased complexity of the expanded dataset. The earlier evaluation used 300 images collected under less complex environmental conditions using an OAK-D Pro camera. In contrast, the current dataset contained 1,008 images with cluttered backgrounds, greater variation in cotton boll sizes, and diverse lighting conditions. It also included 336 images from each of three cameras with different resolutions. The resulting variation in image perspective, resolution, and image quality increased the difficulty of the cotton boll detection task.}

\textcolor{black}{Although newer models such as YOLOv10 and YOLOv11 have demonstrated strong mAP performance on the COCO dataset~\cite{Jocher_Ultralytics_YOLO_2023}, the YOLOv9-based models performed better for the cotton boll detection task evaluated in this study. The performance of YOLOv9-s and GELAN-s may be associated with the Generalized Efficient Layer Aggregation Network architecture, which was designed to improve parameter utilization, and, for YOLOv9-s, with Programmable Gradient Information, which preserves reliable gradient information during training~\cite{wang2024yolov9}. However, these results do not indicate that the models are universally superior to newer YOLO models. Instead, they show that YOLOv9-s and GELAN-s achieved a favorable balance among accuracy, recall, and inference speed for the present dataset.}

\textcolor{black}{The average test mAP@0.5 of 84.5\% obtained by GELAN-s was lower than the 92.75\% mAP reported for MRF-YOLO on small unopened cotton bolls~\cite{liu2023small}. Other cotton studies reported an accuracy of 88\% using UAV imagery and region-growing segmentation~\cite{yeom2018automated} and a detection accuracy of 87.4\% using a small-target YOLO framework with UAV imagery~\cite{zhang2024yolo}. These numerical values are not directly interchangeable because the studies differed in boll maturity, image-acquisition perspective, target size, dataset composition, and performance metrics. In particular, the present study evaluated seed cotton under cluttered ground-level field conditions using images from three cameras and assessed the selected models across five randomized dataset splits. Thus, its contribution is not a claim of universally higher detection accuracy, but a deployment-oriented comparison of multiple contemporary YOLO and GELAN variants based on both accuracy and inference speed. The strong performance of GELAN-s across the randomized splits further suggests that model selection based on repeated dataset partitions can provide a more stable basis for downstream deployment than selection from a single split.}

\subsection{\textcolor{black}{Cotton Boll Segmentation Model Performance}}

\textcolor{black}{Among the direct YOLO-based segmentation models, YOLOv12-m-seg was the only model that exceeded the reference values for both normalized mAP@0.5 and FPS (Figure~\ref{figure:9}). Its performance across the primary and randomized datasets and its accuracy--speed trade-off supported its selection as the direct segmentation candidate. The model was then compared with the detection-prompted segmentation approach based on GELAN-s and the SAM models.}

\textcolor{black}{For the detection-prompted approach, SAM segmentation using direct bounding-box prompts performed better than the clipped-patch approach. The clipped-patch results included portions of cotton leaves and background elements such as clouds, indicating that the contextual information from the original image was not sufficiently preserved when individual cotton bolls were presented as small image patches. The use of bounding-box prompts is consistent with the zero-shot segmentation capability reported for SAM~\cite{kirillov2023segment} and its large-scale pretraining using the SA-1B dataset~\cite{kirillov2023segmentdataset}. Bounding-box prompts provided the spatial information used by SAM to localize and segment the cotton boll regions.}

\textcolor{black}{A closely related study combined YOLO detection with SAM segmentation for cotton-yield estimation from UAV imagery~\cite{reddy2024cotton}. That study reported mAP@0.5 values of 85.7\% and 83.3\% and IoU values of 68.5\% and 68.3\% for YOLOv7~+~SAM and YOLOv8~+~SAM, respectively, and obtained an $R^2$ of 0.913 between the YOLOv8~+~SAM output and measured yield. The present study differs in both objective and acquisition geometry: it uses ground-level multi-camera imagery and requires instance-level cotton boll localization for robotic picking rather than row-level yield prediction from UAV imagery. Nevertheless, both studies show that YOLO-generated spatial information can guide SAM in cotton applications. The present results additionally demonstrate that the performance of such a two-stage pipeline is constrained by the upstream detector because cotton bolls missed by GELAN-s are never supplied to SAMv2.1 Tiny for segmentation.}

\textcolor{black}{The masks produced by the SAMv2.1 Tiny, Small, Base+, and Large variants were largely similar in the representative images. Their segmented areas showed relatively small differences that were not strictly monotonic across the three weather conditions (Table~\ref{table:areanalysis}). Because statistical significance was not evaluated, these results represent only a descriptive comparison among the variants. Given the preference for lightweight models in near real-time robotic operation, SAMv2.1 Tiny was retained for comparison with YOLOv12-m-seg.}

\textcolor{black}{The representative FastSAM-s outputs included portions of the sky, leaves, branches, and stems in the cotton segmentation masks, while FastSAM-x reduced the sky-related segmentation errors but continued to include other plant structures~\cite{zhao2023fast}. Grounded-SAM with RAM produced better segmentation results for the sunny and partially cloudy images but failed to distinguish individual cotton bolls in the cloudy example~\cite{ren2024grounded,kirillov2023segment,liu2023grounding}. Based on these observations, GELAN-s + SAMv2.1 Tiny was retained as the detection-prompted segmentation approach for comparison with YOLOv12-m-seg. These observations are limited to the representative images shown in Figure~\ref{figure:14}.}

\textcolor{black}{The sensitivity of FastSAM and Grounded-SAM to background structures and changing illumination is consistent with earlier cotton-vision studies. Singh et al.~\cite{singh2021image} reported that a chromatic-aberration method achieved an identification rate of 91.05\% under natural illumination but remained susceptible to light reflections and overlapping bolls. Similarly, the U-Net-based approach evaluated by Nagarajan et al.~\cite{nagarajan2023cotton} was developed specifically to distinguish cotton from sky interference. These comparisons indicate that background confusion is not unique to foundation-model segmentation; it remains a persistent challenge across traditional image processing, task-specific convolutional networks, and prompt-based segmentation. Unlike reported validation accuracy or identification rate, the present qualitative observations do not provide a directly comparable numerical benchmark, but they reveal analogous failure modes under field conditions.}

\textcolor{black}{The final comparison showed that GELAN-s + SAMv2.1 Tiny segmented additional back-row cotton bolls in some representative images, although this observation did not independently demonstrate greater robustness (Figure~\ref{figure:15}). In the complete test-set evaluation, YOLOv12-m-seg produced less-negative count-based $R^2$ values than GELAN-s at confidence levels of 80\% and 50\%. Nevertheless, all count-based $R^2$ values were negative, indicating poor absolute agreement with the reference cotton boll counts. Lowering the confidence level slightly improved the agreement for both models, but substantial count differences remained.}

\textcolor{black}{The poor count correspondence contrasts with results from methods designed specifically for cotton boll counting. MRF-YOLO achieved a count $R^2$ of 0.92 for small unopened bolls~\cite{liu2023small}, the UAV-based small-target YOLO approach reported an $R^2$ of 0.86~\cite{zhang2024yolo}, and the image-processing method of Sun et al.~\cite{sun2019image} achieved a boll-count accuracy of 83\%. These outcomes are not direct benchmarks for the present models because they involve different boll stages, viewing geometries, reference annotations, and counting metrics. However, they show that reliable enumeration generally requires a counting-oriented evaluation and explicit treatment of small, overlapping, or densely clustered targets. The negative count-based $R^2$ values in the present study therefore indicate that models selected for segmentation accuracy and robotic localization should not automatically be assumed to provide reliable boll counts.}

\textcolor{black}{In the segmented-area comparison, the automatically annotated masks achieved an $R^2$ value of 0.972 relative to the manual annotations, while YOLOv12-m-seg and GELAN-s + SAMv2.1 Tiny achieved values of 0.966 and 0.860, respectively (Figure~\ref{figure:17}). The lower correspondence of GELAN-s + SAMv2.1 Tiny was associated with cotton bolls that were not detected by GELAN-s and therefore were not passed to SAMv2.1 Tiny for segmentation. In addition, YOLOv12-m-seg required 20.4~ms per image, whereas GELAN-s + SAMv2.1 Tiny required approximately 60~ms per image. Based on the cotton boll count comparison, segmented-area agreement, and inference-speed evaluation, YOLOv12-m-seg was selected for field testing.}

\textcolor{black}{The strong image-level area correspondence is broadly consistent with the use of segmented cotton pixels as an aggregate indicator in the UAV-based YOLO--SAM study, which reported an $R^2$ of 0.913 for yield prediction~\cite{reddy2024cotton}. However, that value describes the relationship between segmented imagery and harvested yield, whereas the present $R^2$ values describe correspondence between predicted and manually annotated pixel areas. The values therefore support the utility of segmented area as a scene-level measure but should not be compared as equivalent accuracy metrics.}

\textcolor{black}{The divergence between the count-based and segmented-area results reflects the different error modes captured by these measures. Count-based evaluation requires each cotton boll to be detected and separated as an individual instance, whereas total segmented area aggregates all cotton pixels within an image. Consequently, missed small or distant bolls may produce substantial counting errors while contributing relatively little to the total segmented area. Similarly, adjacent or overlapping bolls may be represented as a single instance while preserving much of their combined area. Oversegmentation of detected bolls could also compensate for area omitted because of missed detections; however, the present aggregate analysis does not establish whether such compensation occurred. Furthermore, the count evaluation used reference counts derived from the automatically annotated masks, whereas the area evaluation used manually annotated masks; therefore, the corresponding $R^2$ values describe different outcomes and are not directly comparable. These findings indicate that strong image-level area correspondence does not necessarily imply reliable instance-level boll counting. This distinction is particularly important for robotic harvesting because each boll must be detected, separated, and localized as an individual picking target.}

\textcolor{black}{Camera resolution was also identified as a factor influencing detection performance. The lower effective resolution of images captured using the RealSense D435i camera was associated with more frequent missed detections and segmentations, particularly under close-range conditions. These missed identifications contributed to the negative count-based $R^2$ values.}

\textcolor{black}{The observed influence of resolution is consistent with previous work emphasizing small-target representation. MRF-YOLO incorporated multi-receptive-field extraction and a dedicated small-target detection layer to reduce feature loss~\cite{liu2023small}, while the SSPD approach combined space-to-depth operations with a small-target detection head for low-resolution UAV imagery~\cite{zhang2024yolo}. Together with the present camera-specific results, these studies indicate that retaining fine spatial information is particularly important when bolls occupy few pixels or appear behind foreground plant structures.}

\subsection{\textcolor{black}{Field Deployment and Robotic Picking Performance}}

\textcolor{black}{Based on the combined evaluation of model accuracy, cotton boll count agreement, segmentation-area agreement, and inference speed, YOLOv12-m-seg was selected for field testing with the robotic cotton-picking system. Its shorter inference time provided an important benefit for implementing the perception model in the physical system for real-time cotton picking.}

\textcolor{black}{Figure~\ref{figure:18} shows the field experiments conducted using the developed autonomous cotton picker. Figure~\ref{figure:19} presents the in-field cotton boll segmentation and picking process using the ZED2i stereo camera and UR5e manipulator. The model supplied localized target coordinates to the robotic system, and segmentation outputs were obtained during picking trials conducted at confidence levels of 80\% and 70\%.}

\textcolor{black}{During the field experiments, YOLOv12-m-seg identified and segmented seed cotton bolls located in the front rows of the canopy at both confidence levels.} \textcolor{black}{These results demonstrate the operational integration of YOLOv12-m-seg within the Cotton-Eye perception system and establish the feasibility of using the model to generate actionable targets under field conditions.}

\textcolor{black}{Gharakhani et al.~\cite{gharakhani2024field} reported that a field-tested robotic cotton harvester using YOLOv4-tiny detected 78.1\% of visible bolls, localized 70.0\% of the detected bolls, picked 83.1\% of the localized bolls, and harvested 55.1\% of the seed cotton that was visible and within the arm workspace. These percentages cannot be directly compared with the 49.0\% and 57.5\% target-acceptance proportions obtained at the 80\% and 70\% confidence levels in the present study. The prior study reported stage-specific outcomes using visible, detected, localized, or reachable bolls as denominators, whereas acceptance in the present system required a detection to satisfy confidence, image-size, distance, and manipulator-workspace criteria simultaneously. Nevertheless, both studies demonstrate that offline model accuracy alone does not determine robotic-harvesting performance; camera geometry, localization reliability, physical reachability, boll occlusion, and picking mechanics also constrain the number of harvestable targets.}

\section{CONCLUSION}

In this study, a vision-\textcolor{black}{guided} cotton boll detection and segmentation framework was developed and validated for robotic cotton-picking system. Multiple state-of-the-art object detection models, including YOLOv8, YOLOv9, YOLOv10, YOLOv11, YOLOv12, and YOLOv13, were trained and evaluated using a dataset of 1,008 \textcolor{black}{manually} annotated field images. The models were assessed based on detection accuracy, robustness across randomized datasets, and suitability for real-time robotic harvesting.

Among the evaluated detection models \textcolor{black}{for cotton bolls}, GELAN-s demonstrated \textcolor{black}{outstanding} performance under strict operational constraints, particularly in balancing mean Average Precision (mAP) and frames per second (FPS). Its consistent \textcolor{black}{high}-performance across five randomized datasets also highlighted its robustness to dataset variability. As a result, GELAN-s was selected as the detection backbone for further segmentation analysis using \textcolor{black}{automatic} detection-\textcolor{black}{prompted} segmentation pipelines based on SAM and SAMv2.1.

In parallel, direct segmentation models, including YOLOv8-seg, YOLOv11-seg, and YOLOv12-seg, were trained and evaluated. Among these, YOLOv12-m-seg consistently outperformed other segmentation \textcolor{black}{models} in terms of mAP–FPS trade-off, segmentation area consistency, and robustness across randomized datasets. 

\textcolor{black}{The count-based regression analysis showed that YOLOv12-m-seg produced less-negative $R^2$ values than GELAN-s; however, the negative values obtained for both models indicated poor absolute agreement with the reference cotton boll counts. Separately, the area-based evaluation against the manually annotated masks showed that YOLOv12-m-seg achieved an $R^2$ of 0.966 for total segmented area.}

A comparative analysis between detection-\textcolor{black}{prompted} segmentation (GELAN-s + SAMv2.1 \textcolor{black}{Tiny}) and direct segmentation (YOLOv12-m-seg) demonstrated that YOLOv12-m-seg detected and segmented a \textcolor{black}{greater} number of cotton bolls more consistently, while maintaining faster inference speed \textcolor{black}{of 20.4 ms}. Although SAMv2.1-based approaches showed competitive segmentation quality, their performance was \textcolor{black}{relatively} constrained by upstream detection recall and higher computational overhead.

Based on comprehensive quantitative evaluations and real-world field tests using our developed cotton-picking robot, YOLOv12-m-seg was \textcolor{black}{finally} selected as the preferred \textcolor{black}{Cotton-Eye} perception model for \textcolor{black}{robotic} cotton-picking. Field tests conducted using a UR5e robotic arm \textcolor{black}{integrated with} a \textcolor{black}{Husky} mobile platform and ZED2i stereo camera confirmed the \textcolor{black}{YOLOv12-m-seg}’s capability to perform reliable real-time cotton boll detection, segmentation, and picking under varying field conditions and confidence \textcolor{black}{level}s.

The key conclusions of this work are summarized as follows:

\begin{itemize}

	\item GELAN-s achieved \textcolor{black}{desired} detection performance under strict operational constraints, including a confidence \textcolor{black}{level} of at least 80\% and either a bounding box width or height greater than 100 pixels, while achieving an mAP of approximately 86\% \textcolor{black}{with an inference speed of 42.3 ms per image (approximately 23.6 FPS)}.
	
	\item Among all segmentation \textcolor{black}{models}, YOLOv12-m-seg achieved the best balance between segmentation accuracy and inference speed, with an mAP of approximately 83.7\% and an average inference time of about 20.4~ms per image.
	
	\item \textcolor{black}{Automatic detection-prompted} segmentation using \textcolor{black}{GELAN-s + SAMv2.1 Tiny required approximately 60~ms between consecutive inferences during laboratory evaluation, while providing more consistent qualitative segmentation performance than FastSAM and Grounded-SAM with RAM across the evaluated field conditions.}

	\item \textcolor{black}{The area-based evaluation showed that YOLOv12-m-seg achieved an $R^2$ value of 0.966 against the manually annotated segmentation masks, compared with 0.860 for GELAN-s + SAMv2.1 Tiny. However, the negative $R^2$ values obtained in the separate count-based analysis indicate that accurate cotton boll counting remains a challenge for both approaches.}
	
	\item Field \textcolor{black}{tests} validated that YOLOv12-m-seg enabled reliable real-time cotton boll detection, segmentation, and picking using \textcolor{black}{our developed} robotic system, confirming its suitability for practical deployment in \textcolor{black}{robotic} cotton harvesting.
	
\end{itemize}

\textcolor{black}{To conclude}, this work demonstrates that direct segmentation using YOLOv12-m-seg offers a robust, efficient, and scalable perception solution for autonomous cotton-picking robots. The proposed framework provides a \textcolor{black}{solid} foundation for future extensions, including adaptive grasp planning, large-scale field deployment, and \textcolor{black}{yield estimation}.

\section{LIMITATION\textcolor{black}{S} OF THE STUDY}\label{sec:limitation}

\begin{figure}[hbt!]
	\centering
	\begin{subfigure}[b]{0.50\textwidth}
		\centering
		\includegraphics[height=5cm]{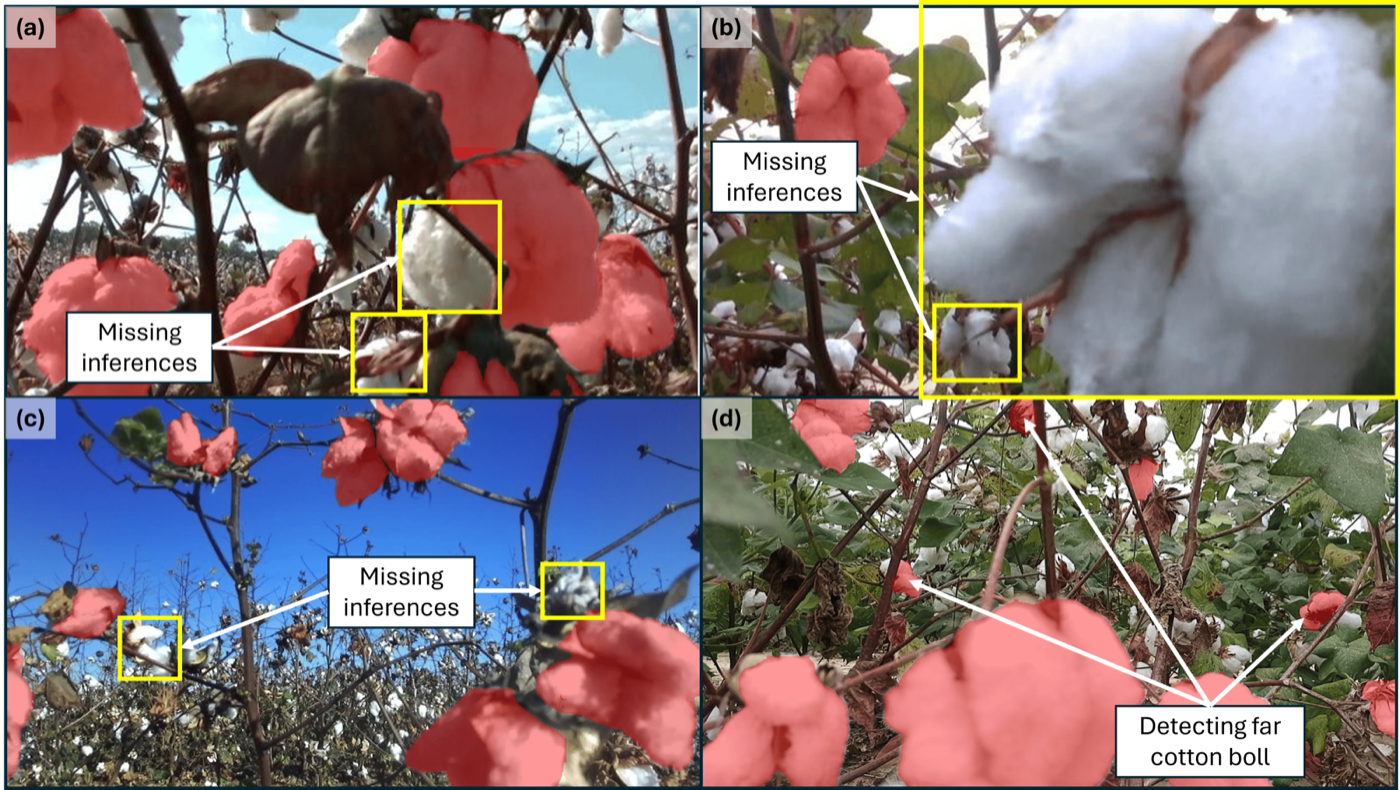}
	\end{subfigure}
	\caption{\textcolor{black}{Representative perception failure cases observed during field evaluation using YOLOv12-m-seg. Panels (a), (b), and (c) show partial or complete segmentation failures involving medium and large cotton bolls. Panel (d) shows the detection of a cotton boll located behind the plant canopy and outside the intended working area.}}
	\label{figure:20}
\end{figure}

Despite the promising results obtained in this study, several limitations were observed during quantitative evaluation and field deployment. These limitations provide important insights for improving the robustness of the cotton boll perception system. \textcolor{black}{The principal limitations concern the reference annotations, dataset composition, segmentation failures, depth ambiguity, crop occlusion, and camera sensing conditions.}

\textcolor{black}{The first limitation concerns the reference masks used for the quantitative evaluation of the segmentation models. The reported segmentation \textcolor{black}{mAP@0.5} values were calculated using the automatically generated segmentation masks rather than the manually annotated masks. The manually annotated masks for the 105 test images were used only as an independent reference for the area based regression analysis. Therefore, the reported segmentation \textcolor{black}{mAP@0.5} values should be interpreted as agreement with the automatically annotated segmentation dataset, whereas the area based analysis provides an independent comparison with the manually annotated ground truth.}

\textcolor{black}{A further limitation concerns the field-test outcome records. The available records retained only the final acceptance or rejection status and did not identify which confidence, bounding-box size, distance, or manipulator-workspace criterion caused an individual rejection. Consequently, the rejected detections could not be retrospectively disaggregated by rejection rule. In addition, the field evaluation did not separately quantify picking attempts and successful cotton removal; therefore, the reported acceptance proportions characterize perception-system actionability rather than end-to-end harvesting effectiveness.}

As shown in Figures~\ref{figure:20}a,~\ref{figure:20}b, and~\ref{figure:20}c, YOLOv12-m-seg occasionally failed to fully segment medium and large cotton bolls, resulting in partial or complete segmentation failures. Such cases were more frequently observed in images captured using the Intel RealSense D435i camera, particularly under close range imaging and lower effective resolution conditions. Reduced spatial resolution and limited contextual information in these scenarios restrict the model's ability to accurately delineate cotton boundaries, especially when fine structural details are required.

Another limitation arises from the composition of the dataset used in this study. Images collected under sunny, partially cloudy, and cloudy conditions were combined into a single dataset without explicit stratification by environmental conditions. This mixed distribution may introduce imbalance across lighting scenarios, potentially affecting the model's ability to generalize uniformly under all field conditions. Additional training data explicitly categorized by weather and illumination conditions could further enhance detection and segmentation robustness.

\textcolor{black}{Natural crop variability also affected perception performance. Leaves, branches, neighboring cotton bolls, and dense canopy structures can partially or completely obscure harvestable bolls. Variations in boll size, opening stage, position, orientation, plant architecture, and canopy density further change the visual appearance of the targets. Background cotton bolls visible through gaps in the canopy can also be detected even when they are outside the intended working area. These conditions increase the difficulty of distinguishing individual cotton bolls and obtaining complete segmentation masks.}

Depth ambiguity and occlusion effects also present challenges for reliable cotton boll localization. As illustrated in Figure~\ref{figure:20}d, the model occasionally detected cotton bolls located behind the plant canopy under the existing filtering constraints. \textcolor{black}{These detections can introduce uncertainty when the perception results are combined with depth measurements and may produce target coordinates outside the intended working area. This limitation highlights the need for improved depth filtering and canopy aware perception to distinguish accessible foreground cotton bolls from background detections.}

Furthermore, sensing constraints such as camera resolution, viewpoint, and field of view significantly influenced detection and segmentation performance. In particular, close proximity images captured by the RealSense D435i camera resulted in more missed detections due to reduced scene context and perspective distortion. These observations emphasize the need for optimized camera placement, improved sensor resolution, or multiple view perception to enhance robustness under field conditions.

\section{FUTURE WORK}\label{sec:futurework}

Future work will focus on further improving the robustness and generalization capability of YOLOv12-m-seg by fine tuning it with additional cotton field images collected under diverse and explicitly balanced environmental conditions. Organizing the dataset according to illumination and weather conditions is expected to improve segmentation consistency and reduce performance variability under challenging lighting scenarios. In addition, hyperparameter optimization and architectural configuration tuning using genetic algorithms will be explored to further enhance model performance.

\textcolor{black}{The expanded dataset will include images representing different canopy densities, cotton boll sizes, opening stages, levels of occlusion, boll orientations, plant architectures, and background conditions. Images collected from additional fields, growing seasons, cultivars, and camera systems will also be incorporated to support independent external evaluation. Automatically generated annotations may be used to accelerate dataset development, followed by human verification to maintain annotation quality.}

Image resolution was identified as a critical factor influencing seed cotton segmentation accuracy and inference speed. Although the OAK-D Pro and ZED2i cameras provide sufficiently high resolution images for cotton boll segmentation, increased resolution leads to higher computational cost and latency, which can affect real time perception performance. Future studies will therefore investigate the relationship among image resolution, segmentation accuracy, and inference time to determine suitable operating parameters for field deployment. Future work will also explore lightweight feature extraction and region of interest filtering before segmentation to reduce unnecessary background processing while maintaining segmentation accuracy. Such preprocessing strategies may further improve inference speed without compromising segmentation quality.

\textcolor{black}{Multiple view perception and temporal tracking will be explored to improve the detection of partially occluded cotton bolls and maintain target identity across consecutive image frames. These approaches may also reduce duplicate detections and improve perception stability when the camera or cotton plants move during field operation.}

\textcolor{black}{Extended perception evaluations will be conducted across different crop growth stages, plant densities, weather conditions, camera viewpoints, and field locations. These evaluations will assess detection, segmentation, depth localization, and inference consistency under conditions that were not represented in the current dataset.}

\textcolor{black}{Future field evaluations will implement stage-specific outcome logging for detection, each rejection criterion, depth localization, picking attempts, successful cotton removal, and cycle time. Recording these outcomes separately will identify the principal constraints on system performance and enable direct comparison with prior robotic cotton-harvesting studies that report detection, localization, and picking as distinct stages.}

\section*{ACKNOWLEDGMENTS}

This work is supported by the Cotton Incorporated (Grant Nos. 23-889 \textcolor{black}{\& 25-582}). The authors also acknowledge support from the Mississippi
Agricultural and Forestry Experiment Station (MAFES) and the Department
of Agricultural and Biological Engineering at Mississippi State University. Any opinions, findings, conclusions, or recommendations expressed in this publication are those of the authors and should not be construed to represent any views of the funding agency and/or the institutions that the authors are affiliated with.

\section*{DATA AND CODE AVAILABILITY}

The image dataset, bounding-box and segmentation annotations, trained
model weights, and implementation code used in this study are publicly
available through the CottonBoll\_Harvest GitHub repository:
\url{https://github.com/imtheva/CottonBoll_Harvest}.

\section*{DECLARATION OF GENERATIVE AI AND AI-ASSISTED TECHNOLOGIES IN THE WRITING PROCESS}

The authors acknowledge the use of ChatGPT \textcolor{black}{(GPT-5.2, OpenAI; accessed January 2026)} during the preparation of this manuscript to enhance its language quality and improve overall readability. All content generated or refined with the assistance of this tool was carefully reviewed, edited, and verified by the authors, who accept full responsibility for the final published version of the article.

\appendix
\section{}

\begin{table}[ht]
	\caption{Five randomized cotton dataset splits (70\% train, 10\% validation, \textcolor{black}{and} 20\% test) used to evaluate model robustness and performance consistency. Each split was generated from the same 1,008-image pool and includes the corresponding number of \textcolor{black}{manually} annotated cotton bolls.}
	
	\label{table:1c}
	\begin{center}       
		\begin{tabular}{|l|l|l|l|l|}
			\hline
			\makecell{\textbf{Dataset}\\\textbf{\textcolor{black}{No.}}} &			
			\textbf{Split} &
			\textbf{Images\# (\%)} &
			\makecell{\textbf{Cotton bolls\#}\\\textbf{(\%)}} \\ \hline
			
			\multirow{4}{*}{Dataset 1} & Train & 705 (70\%) & 16,787 (70.0\%) \\ \cline{2-4}
			& Val & 100 (10\%) & 2,290 (9.6\%) \\ \cline{2-4}
			& Test & 203 (20\%)& 4,883 (20.4\%) \\ \cline{2-4}
			& \textbf{Total}& \textbf{1,008 (100\%)} & \textbf{23,960 (100\%)} \\ \hline
			\multirow{4}{*}{Dataset 2} & Train & 705 (70\%) & 16,606 (69.3\%) \\ \cline{2-4}
			& Val & 100 (10\%) & 2,396 (10.0\%) \\ \cline{2-4}
			& Test & 203 (20\%)& 4,958 (20.7\%) \\ \cline{2-4}
			& \textbf{Total}& \textbf{1,008 (100\%)} & \textbf{23,960 (100\%)} \\ \hline
			\multirow{4}{*}{Dataset 3} & Train & 705 (70\%) & 16,933 (70.7\%) \\ \cline{2-4}
			& Val & 100 (10\%) & 2,295 (9.6\%) \\ \cline{2-4}
			& Test & 203 (20\%)& 4,732 (19.7\%) \\ \cline{2-4}
			& \textbf{Total}& \textbf{1,008 (100\%)} & \textbf{23,960 (100\%)} \\ \hline
			\multirow{4}{*}{Dataset 4} & Train & 705 (70\%) & 16,595 (69.3\%) \\ \cline{2-4}
			& Val & 100 (10\%) & 2,632 (11.0\%) \\ \cline{2-4}
			& Test & 203 (20\%)& 4,733 (19.7\%) \\ \cline{2-4}
			& \textbf{Total}& \textbf{1,008 (100\%)} & \textbf{23,960 (100\%)} \\ \hline
			\multirow{4}{*}{Dataset 5} & Train & 705 (70\%) & 16,933 (70.7\%) \\ \cline{2-4}
			& Val & 100 (10\%) & 2,412 (10.1\%) \\ \cline{2-4}
			& Test & 203 (20\%)& 4,615 (19.2\%) \\ \cline{2-4}
			& \textbf{Total}& \textbf{1,008 (100\%)} & \textbf{23,960 (100\%)} \\ \hline

		\end{tabular}
	\end{center}
\end{table}

\begin{table*}[hbt!]
	\caption{Training and validation results of the YOLO models for cotton boll detection. IoU refers to Intersection-over-Union. Total \textcolor{black}{number of epochs is set to 300.}} 
	\label{table:8}
	\begin{center}       
		
		\begin{tabular}{|l|l|l|l|l|l|l|}
			\hline
			\textbf{YOLO Model}& \textbf{Dataset No.} & \textbf{mAP (\%) (IoU=0.5)} & \textbf{Precision (\%)}& \textbf{Recall (\%)}& \textbf{F1-Score \textcolor{black}{(\%)}} & \textbf{Epochs (Best \textcolor{black}{Weight})}\\\hline
			\multirow{5}{*}{GELAN-s} & Dataset 1 & 86.0 & 82.6 & 75.8 & 79.1 & 299 (299) \\ \cline{2-7}
			& Dataset 2 & 85.8 & 80.2 & 75.6 & 77.8 & 299 (299) \\ \cline{2-7}
			& Dataset 3 & 84.5 & 81.1 & 75.3 & 78.1 & 272 (172) \\ \cline{2-7}
			& Dataset 4 & 85.0 & 79.8 & 76.6 & 78.2 & 278 (178) \\ \cline{2-7}
			& Dataset 5 & 85.4 & 80.5 & 75.5 & 77.9 & 299 (299) \\ \hline
			
			\multirow{5}{*}{YOLOv10-l} & Dataset 1 & 82.7 & 80.0 & 72.5 & 76.1 & 144 (44) \\ \cline{2-7}
			& Dataset 2 & 82.6 & 77.0 & 74.0 & 75.5 & 148 (48) \\ \cline{2-7}
			& Dataset 3 & 82.3 & 78.7 & 74.2 & 76.4 & 150 (50) \\ \cline{2-7}
			& Dataset 4 & 81.6 & 76.7 & 75.3 & 76.0 & 182 (82) \\ \cline{2-7}
			& Dataset 5 & 82.5 & 77.6 & 73.5 & 75.5 & 148 (48) \\ \hline
			
			\multirow{5}{*}{YOLOv12-s} & Dataset 1 & 84.3 & 79.1 & 76.7 & 77.9 & 189 (89) \\ \cline{2-7}
			& Dataset 2 & 84.1 & 78.2 & 75.0 & 76.6 & 142 (42) \\ \cline{2-7}
			& Dataset 3 & 82.9 & 77.4 & 74.0 & 75.7 & 142 (42) \\ \cline{2-7}
			& Dataset 4 & 82.6 & 81.9 & 72.4 & 76.9 & 144 (44) \\ \cline{2-7}
			& Dataset 5 & 83.5 & 80.8 & 73.3 & 76.9 & 161 (61) \\ \hline
			
			\multirow{5}{*}{YOLOv12-m} & Dataset 1 & 84.5 & 76.0 & 77.8 & 76.9 & 159 (59) \\ \cline{2-7}
			& Dataset 2 & 83.7 & 76.8 & 76.8 & 76.8 & 142 (42) \\ \cline{2-7}
			& Dataset 3 & 83.3 & 76.6 & 76.9 & 76.7 & 189 (89) \\ \cline{2-7}
			& Dataset 4 & 83.2 & 79.2 & 75.6 & 77.4 & 200 (100) \\ \cline{2-7}
			& Dataset 5 & 83.0 & 77.9 & 74.3 & 76.1 & 175 (75) \\ \hline

		\end{tabular}
	\end{center}
\end{table*}

\begin{table*}[hbt!]
	\caption{Testing results of the YOLO models for cotton boll detection. IoU refers to Intersection-over-Union. Total \textcolor{black}{number of epochs is set to 300.} \textcolor{black}{FPS refers to frames per second.}} 
	\label{table:9}
	\begin{center}       
		
		\begin{tabular}{|l|l|l|l|l|l|l|l|}
			\hline
			\makecell{\textbf{YOLO}\\\textbf{Model}} &
			\makecell{\textbf{Dataset}\\\textbf{No.}} &
			\makecell{\textbf{mAP (\%)}\\\textbf{(IoU=0.5)}} &
			\makecell{\textbf{Precision}\\\textbf{(\%)}} &
			\makecell{\textbf{Recall}\\\textbf{(\%)}} &
			\textbf{F1-Score \textcolor{black}{(\%)}} &
			\makecell{\textbf{Inference Speed}\\\textbf{Per Image (ms)}} &
			\textbf{FPS} \\\hline
			
			\multirow{5}{*}{GELAN-s} & Dataset 1 & 84.8 & 79.6 & 75.7 & 77.6 & 23.9 & 41.8 \\ \cline{2-8}
			& Dataset 2 & 83.4 & 76.9 & 76.2 & 76.5 & 27.5 & 36.4 \\ \cline{2-8}
			& Dataset 3 & 84.8 & 79.2 & 76.8 & 78.0 & 29.7 & 33.7 \\ \cline{2-8}
			& Dataset 4 & 85.2 & 80.4 & 76.2 & 78.2 & 25.3 & 39.5 \\ \cline{2-8}
			& Dataset 5 & 84.4 & 80.0 & 75.6 & 77.7 & 27.0 & 37.0 \\ \hline
			
			\multirow{5}{*}{YOLOv10-l} & Dataset 1 & 82.4 & 77.9 & 73.1 & 75.4 & 28.6 & 35.0 \\ \cline{2-8}
			& Dataset 2 & 81.0 & 76.3 & 72.6 & 74.4 & 25.1 & 39.8 \\ \cline{2-8}
			& Dataset 3 & 82.9 & 77.4 & 75.3 & 76.3 & 33.6 & 29.8 \\ \cline{2-8}
			& Dataset 4 & 82.1 & 77.5 & 76.5 & 76.0 & 28.4 & 35.2 \\ \cline{2-8}
			& Dataset 5 & 82.5 & 76.9 & 74.9 & 75.9 & 28.3 & 35.3 \\ \hline
			
			\multirow{5}{*}{YOLOv12-s} & Dataset 1 & 83.1 & 79.0 & 74.4 & 76.6 & 24.8 & 40.3 \\ \cline{2-8}
			& Dataset 2 & 81.6 & 76.7 & 73.3 & 75.0 & 26.2 & 38.2 \\ \cline{2-8}
			& Dataset 3 & 82.3 & 77.7 & 72.7 & 75.1 & 28.9 & 34.6 \\ \cline{2-8}
			& Dataset 4 & 84.1 & 80.0 & 74.0 & 76.9 & 20.5 & 48.8 \\ \cline{2-8}
			& Dataset 5 & 82.6 & 80.0 & 72.5 & 76.1 & 24.5 & 40.8 \\ \hline
			
			\multirow{5}{*}{YOLOv12-m} & Dataset 1 & 82.5 & 77.9 & 74.3 & 76.1 & 27.3 & 36.6 \\ \cline{2-8}
			& Dataset 2 & 82.4 & 76.1 & 74.7 & 75.4 & 28.5 & 35.1 \\ \cline{2-8}
			& Dataset 3 & 82.9 & 76.3 & 76.1 & 76.2 & 32.4 & 30.9 \\ \cline{2-8}
			& Dataset 4 & 84.0 & 78.7 & 76.5 & 77.6 & 27.7 & 36.1 \\ \cline{2-8}
			& Dataset 5 & 82.7 & 78.2 & 74.6 & 76.4 & 31.4 & 31.8 \\ \hline

		\end{tabular}
	\end{center}
\end{table*}

\begin{figure}[hbt!]
	\centering
	\begin{subfigure}[b]{0.49\textwidth}
		\centering
		\includegraphics[height=8cm]{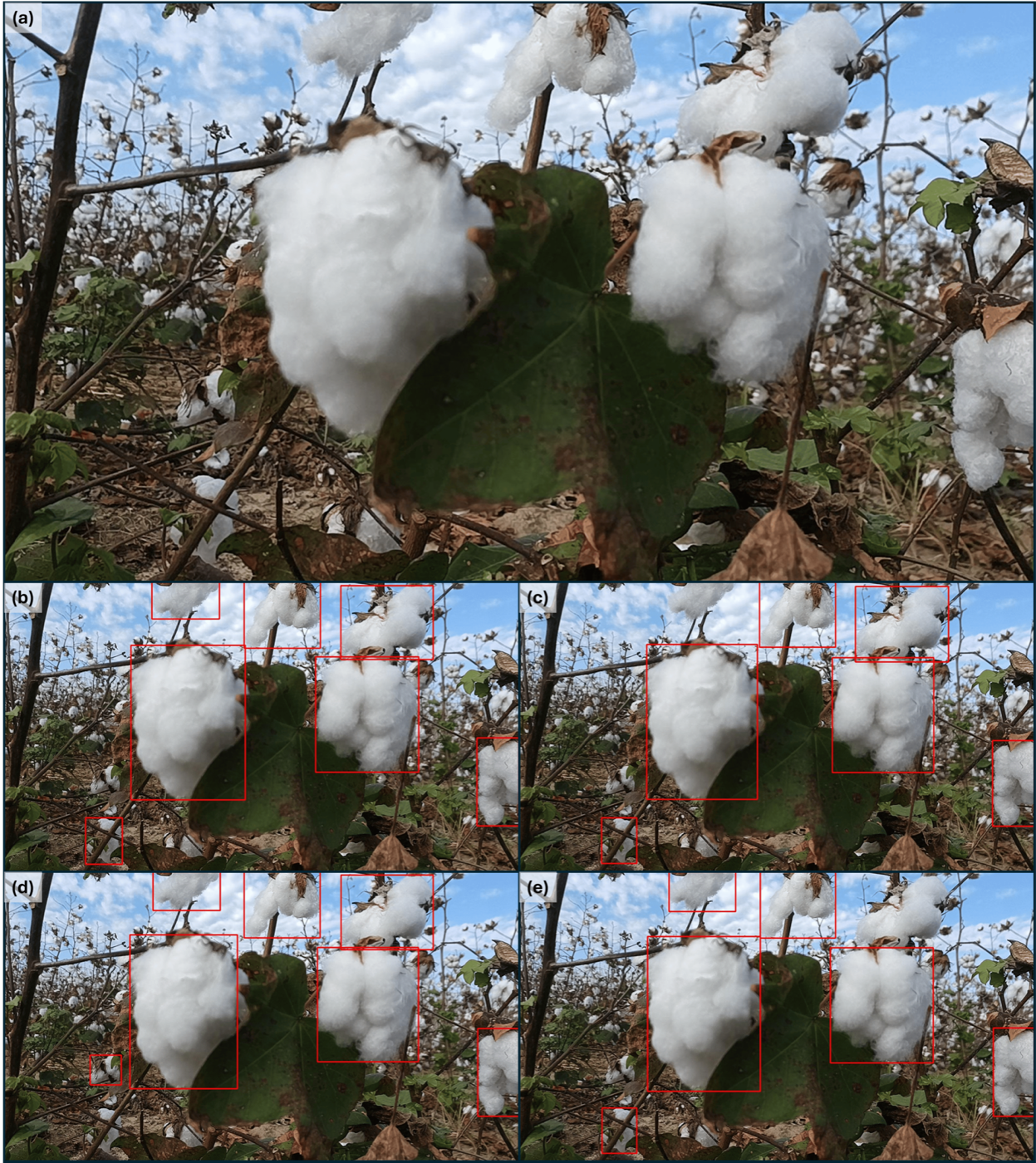}
	\end{subfigure}
	
	\caption{\textcolor{black}{Qualitative cotton boll detection results produced by the four candidate models using a confidence level of at least 80\% and a minimum bounding box width or height greater than 100 pixels. Panel (a) shows the original test image, while panels (b)--(e) show the corresponding results produced by GELAN-s, YOLOv10-l, YOLOv12-s, and YOLOv12-m, respectively.}}
	\label{figure:8}    
\end{figure}

\begin{table*}[hbt!]
	\caption{\textcolor{black}{Training and validation results of YOLOv12-m-seg for cotton boll segmentation across five randomized dataset splits. IoU refers to Intersection over Union. The maximum number of training epochs was set to 300.}}
	\label{table:10}
	\begin{center}       
		
		\begin{tabular}{|l|l|l|l|l|l|l|}
			\hline
			\textbf{YOLO Model}& \textbf{Dataset No.} & \textbf{mAP (\%) (IoU=0.5)} & \textbf{Precision (\%)}& \textbf{Recall (\%)}& \textbf{F1-Score \textcolor{black}{(\%)}} & \textbf{Epochs (Best \textcolor{black}{Weight})}\\\hline
			\multirow{5}{*}{YOLOv12-m-seg} & Dataset 1 & 83.9 & 81.1 & 77.6 & 79.3 & 300 (300) \\ \cline{2-7}
			& Dataset 2 & 84.7 & 81.9 & 76.5 & 79.1 & 221 (121) \\ \cline{2-7}
			& Dataset 3 & 83.4 & 81.2 & 75.2 & 78.1 & 246 (146) \\ \cline{2-7}
			& Dataset 4 & 84.5 & 80.1 & 78.7 & 79.4 & 300 (300) \\ \cline{2-7}
			& Dataset 5 & 84.6 & 80.9 & 78.2 & 79.5 & 300 (300) \\ \hline

		\end{tabular}
	\end{center}
\end{table*}

\begin{table*}[hbt!]
	\caption{\textcolor{black}{Testing results of YOLOv12-m-seg for cotton boll segmentation across five randomized dataset splits. IoU refers to Intersection over Union. FPS refers to frames per second.}}
	\label{table:11}
	\begin{center}       
		
		\begin{tabular}{|l|l|l|l|l|l|l|l|}
			\hline
			\makecell{\textbf{YOLO}\\\textbf{Model}} &
			\makecell{\textbf{Dataset}\\\textbf{No.}} &
			\makecell{\textbf{mAP (\%)}\\\textbf{(IoU=0.5)}} &
			\makecell{\textbf{Precision}\\\textbf{(\%)}} &
			\makecell{\textbf{Recall}\\\textbf{(\%)}} &
			\textbf{F1-Score \textcolor{black}{(\%)}} &
			\makecell{\textbf{Inference Speed}\\\textbf{Per Image (ms)}} &
			\textbf{FPS} \\\hline
			
			\multirow{5}{*}{YOLOv12-m-seg} & Dataset 1 & 83.8 & 82.5 & 75.9 & 79.1 & 15.8 & 63.3 \\ \cline{2-8}			
			& Dataset 2 & 84.2 & 80.4 & 76.3 & 78.3 & 15.9 & 62.9 \\ \cline{2-8}			
			& Dataset 3 & 84.1 & 82.2 & 75.4 & 78.7 & 16.2 & 61.7 \\ \cline{2-8}
			
			& Dataset 4 & 84.7 & 81.7 & 78.4 & 80.0 & 16.9 & 59.2 \\ \cline{2-8}
			
			& Dataset 5 & 83.4 & 81.3 & 75.2 & 78.1 & 15.9 & 62.9 \\ \hline

		\end{tabular}
	\end{center}
\end{table*}

\begin{figure}[hbt!]
	\centering
	\begin{subfigure}[b]{0.49\textwidth}
		\centering
		\includegraphics[height=7cm]{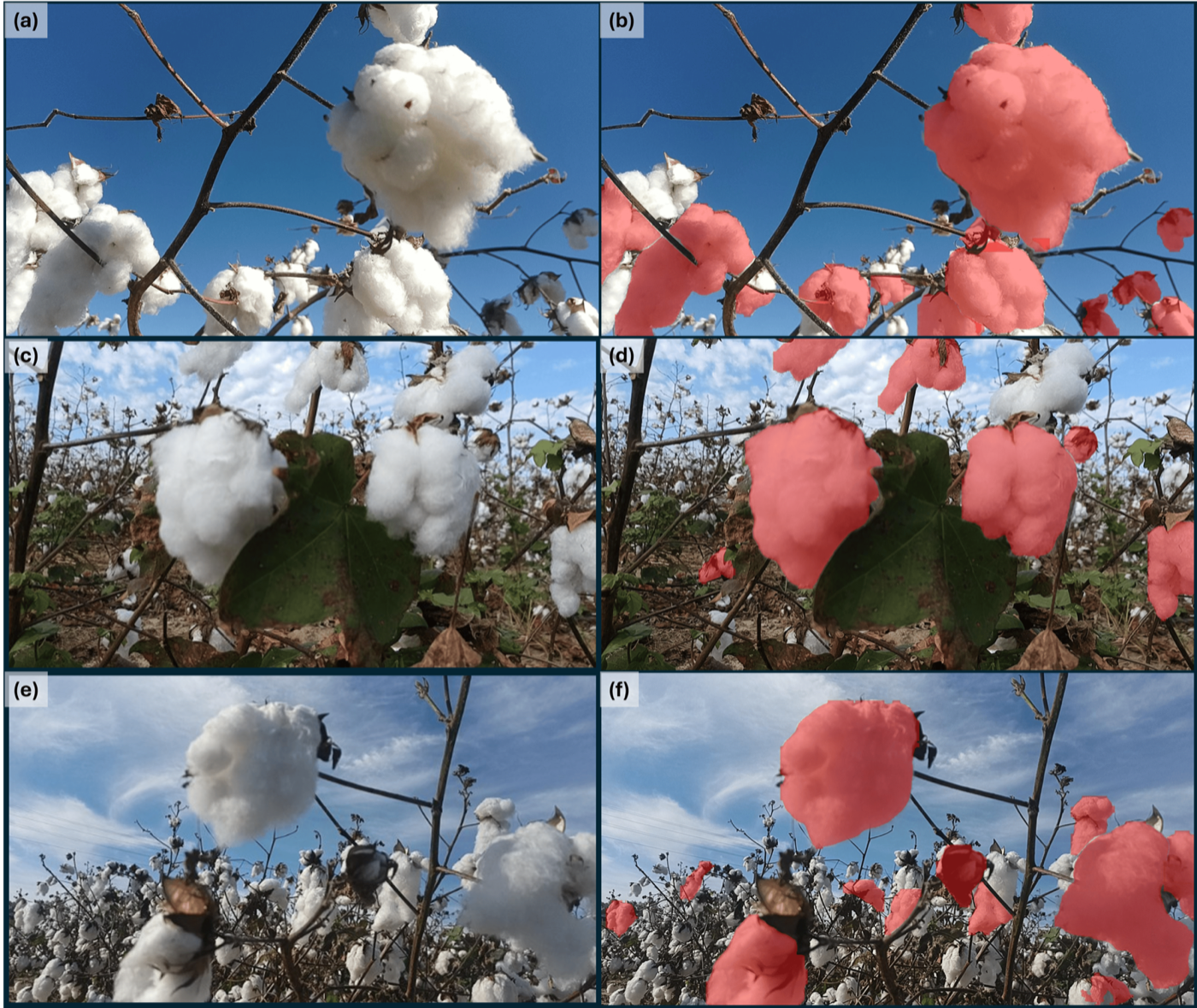}
	\end{subfigure}
	\caption{\textcolor{black}{Representative cotton boll segmentation results produced by YOLOv12-m-seg under different weather conditions. Panels (a), (c), and (e) show the original sunny, partially cloudy, and cloudy field images, respectively. Panels (b), (d), and (f) show the corresponding segmentation outputs.}}
	\label{figure:10}    
\end{figure}

\begin{figure}[hbt!]
	\centering
	\begin{subfigure}[b]{0.49\textwidth}
		\centering
		\includegraphics[height=7cm]{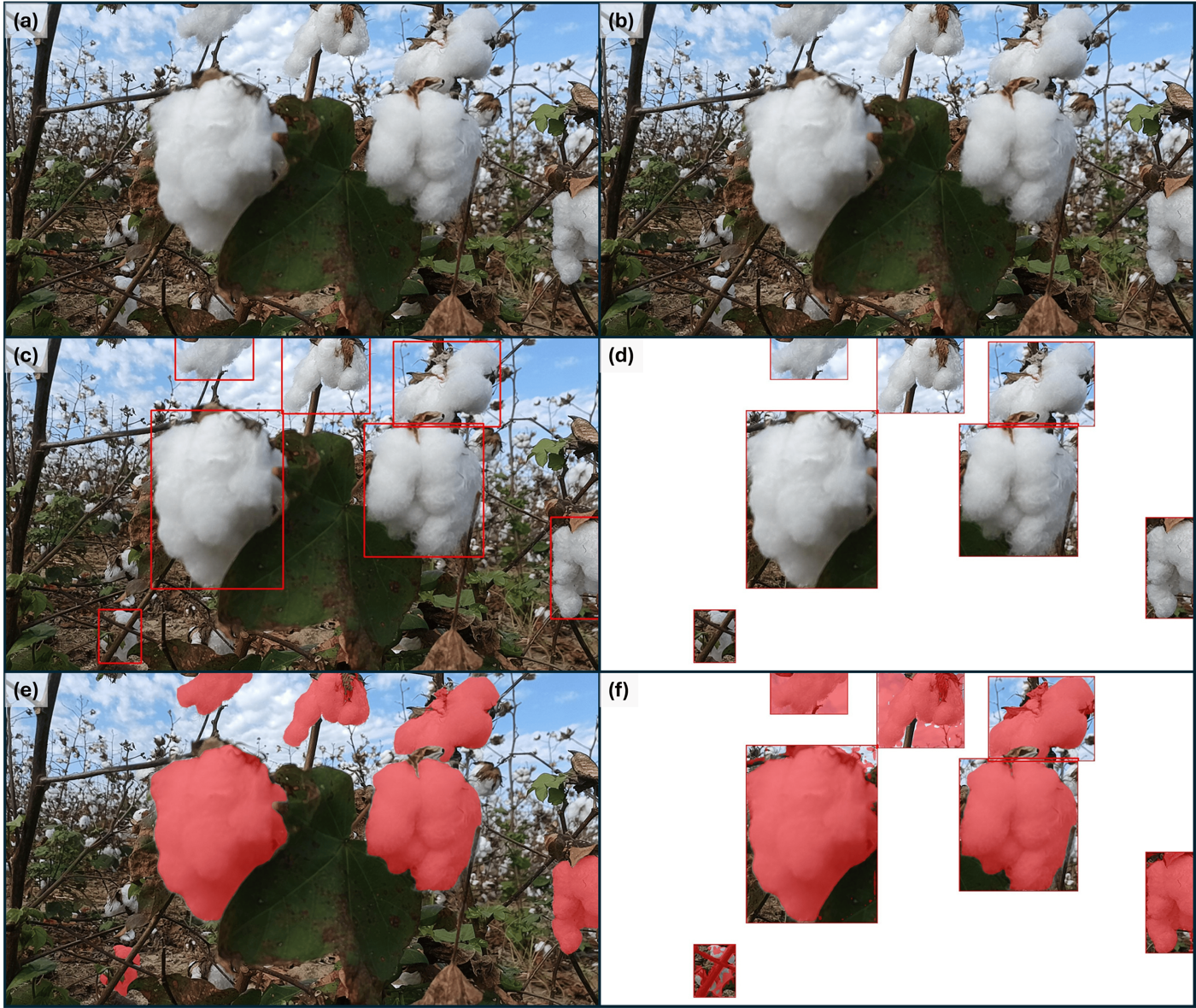}
	\end{subfigure}
	\caption{Comparison of two SAM-based segmentation approaches: (1) Bounding box-based SAM segmentation - (a) Original image, (c) Detected cotton bolls using GELAN-s \textcolor{black}{(best-performing detection model)}, and (e) Segmented cotton bolls using SAM. (2) Image processing-based SAM segmentation - (b) Original image, (d) Processed image retaining only \textcolor{black}{clipped patches of} cotton bolls after GELAN-s detection, and (f) Segmented cotton bolls using SAM.}
	\label{figure:11}    
\end{figure}

\begin{figure}[hbt!]
	\centering
	\begin{subfigure}[b]{0.49\textwidth}
		\centering
		\includegraphics[height=8cm]{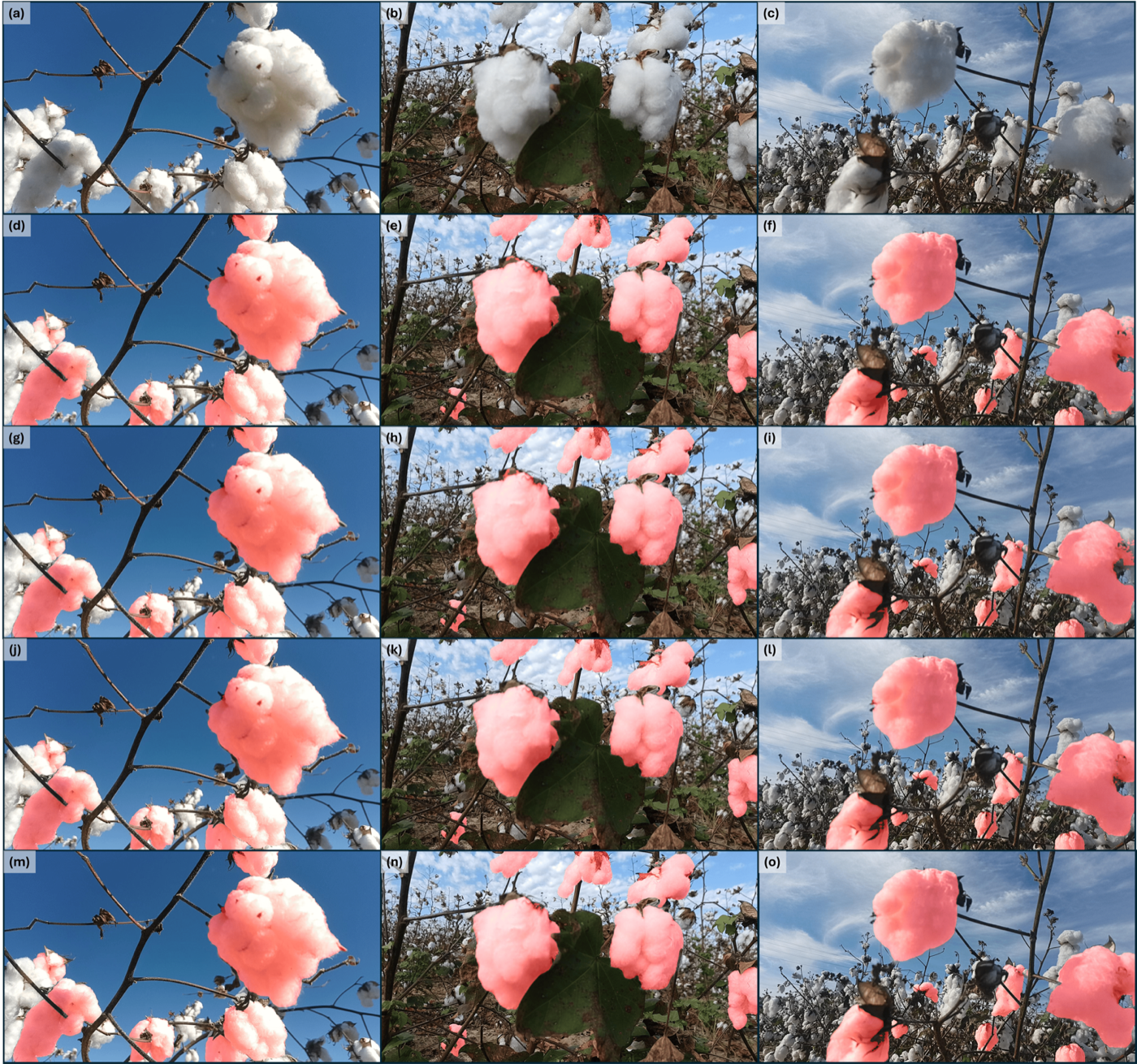}
	\end{subfigure}
	\caption{Cotton boll segmentation performance using GELAN-s + SAMv2.1 variants under different weather conditions: (a) sunny, (b) partially cloudy, and (c) cloudy field conditions. Segmentation results were generated using (d)--(f) SAMv2.1 Tiny, (g)--(i) SAMv2.1 Small, (j)--(l) SAMv2.1 Base+, and (m)--(o) SAMv2.1 Large. \textcolor{black}{Segmentation was automatically prompted using bounding box coordinates inferred by GELAN-s.}}
	\label{figure:13}    
\end{figure}

\begin{figure}[hbt!]
	\centering
	\begin{subfigure}[b]{0.50\textwidth}
		\centering
		\includegraphics[height=8cm]{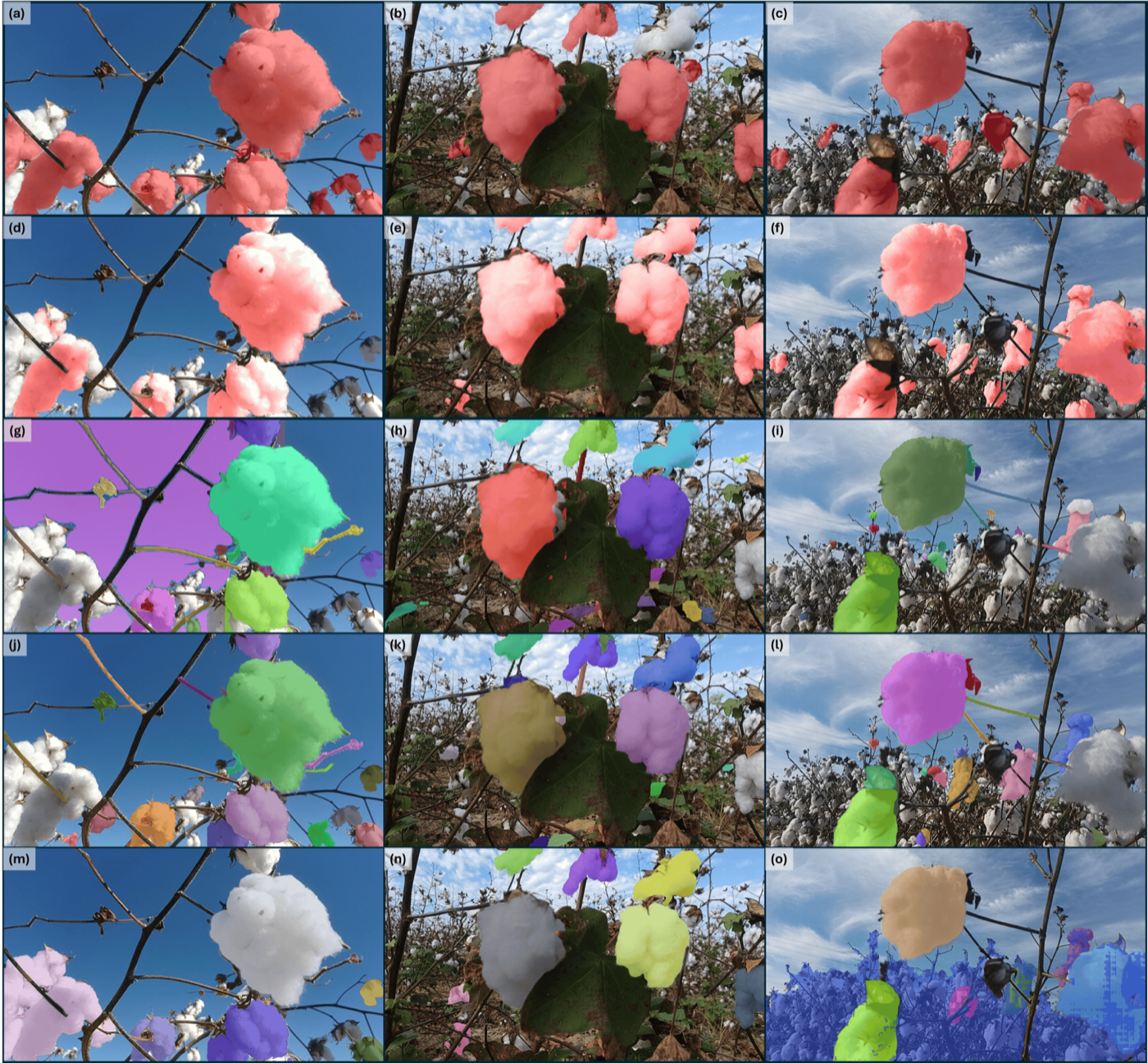}
	\end{subfigure}
	\caption{Qualitative comparison of segmentation performance across multiple state-of-the-art models using three representative field images. Results include GELAN-s + SAM and GELAN-s + SAMv2.1 Tiny, FastSAM-s and FastSAM-x using the text prompt ``cotton bolls,'' and Grounded-SAM with RAM. Panels (a–c) show SAM results; (d–f) SAMv2.1 Tiny results; (g–i) FastSAM-s results; (j–l) FastSAM-x results; and (m–o) Grounded-SAM with RAM results.}
	
	\label{figure:14}    
\end{figure}

\bibliographystyle{elsarticle-num} 
\bibliography{referencessecond}

\end{document}